\pdfoutput=1

\documentclass[11pt]{article}

\usepackage[preprint]{acl}

\usepackage{times}
\usepackage{latexsym}

\usepackage[T1]{fontenc}

\usepackage[utf8]{inputenc}

\usepackage{microtype}

\usepackage{inconsolata}

\usepackage{graphicx}

\usepackage{amsmath}
\usepackage{todonotes}
\newtheorem{hyp}{Scenario}

\title{Beyond Random Sampling: Efficient Language Model Pretraining via Curriculum Learning}

\author{
 \textbf{Yang Zhang\textsuperscript{1}$^\dagger$},
 \textbf{Amr Mohamed\textsuperscript{1,2}},
 \textbf{Hadi Abdine\textsuperscript{2}},
\\
 \textbf{Guokan Shang\textsuperscript{2}$^\dagger$},
 \textbf{Michalis Vazirgiannis\textsuperscript{1,2}}
\\
\\
 \textsuperscript{1}Ecole Polytechnique,
 \textsuperscript{2}MBZUAI
\\
 \small{
   $^\dagger$Correspondence: \texttt{yang.zhang@polytechnique.edu, guokan.shang@mbzuai.ac.ae}
 }
}

\begin{document}
\maketitle
\begin{abstract}
Curriculum learning—organizing training data from easy to hard—has improved efficiency across machine learning domains, yet remains underexplored for language model pretraining. We present the first systematic investigation of curriculum learning in LLM pretraining, with over 200 models trained on up to 100B tokens across three strategies: vanilla curriculum learning, pacing-based sampling, and interleaved curricula, guided by six difficulty metrics spanning linguistic and information-theoretic properties. We evaluate performance on eight benchmarks under three realistic scenarios: limited data, unlimited data, and continual training. Our experiments show that curriculum learning consistently accelerates convergence in early and mid-training phases, reducing training steps by 18-45\% to reach baseline performance. When applied as a warmup strategy before standard random sampling, curriculum learning yields sustained improvements up to 3.5\%. We identify compression ratio, lexical diversity (MTLD), and readability (Flesch Reading Ease) as the most effective difficulty signals. Our findings demonstrate that data ordering—orthogonal to existing data selection methods—provides a practical mechanism for more efficient LLM pretraining.
\end{abstract}

\section{Introduction}

Large language models (LLMs) have achieved remarkable progress across diverse NLP tasks \cite{kaplan2020scaling, achiam2023gpt, anil2023palm}. Yet, this progress comes at enormous computational and data costs, driving increasing interest in methods that make pretraining more efficient. Many efforts focus on improving data quality through filtering \cite{tirumala2023d4, sorscher2022beyond, longpre2024pretrainer} and optimizing data mixtures \cite{xie2023doremi, sachdeva2024train}. However, the ordering of training data remains largely overlooked. Unlike humans, who learn progressively from simple to complex concepts, LLMs are typically trained on randomly sampled data — a process that ignores the possible benefits of structured learning.

Among potential approaches, curriculum learning (CL) \cite{10.1145/1553374.1553380} offers a promising framework: by presenting training data in a structured progression from easy to hard examples, it mimics human learning process and has been shown to improve optimization and generalization in various domains like computer vision \cite{kumar2010self, sinha2020curriculum}. This structured learning approach aims to achieve two important benefits: (i) faster convergence, and (ii) improved best model performance \cite{10.1145/1553374.1553380}.


Despite its intuitive motivation, curriculum learning remains underexplored in the context of LLM pretraining, due to key challenges in defining difficulty measures and designing effective curriculum schedulers for large-scale training \cite{soviany2022curriculum}. In this work, we seek to answer the question: \textit{How can we build effective curriculum learning strategies for LLM pretraining?}

\begin{figure*}[t]
  \centering
  \includegraphics[width=\linewidth]{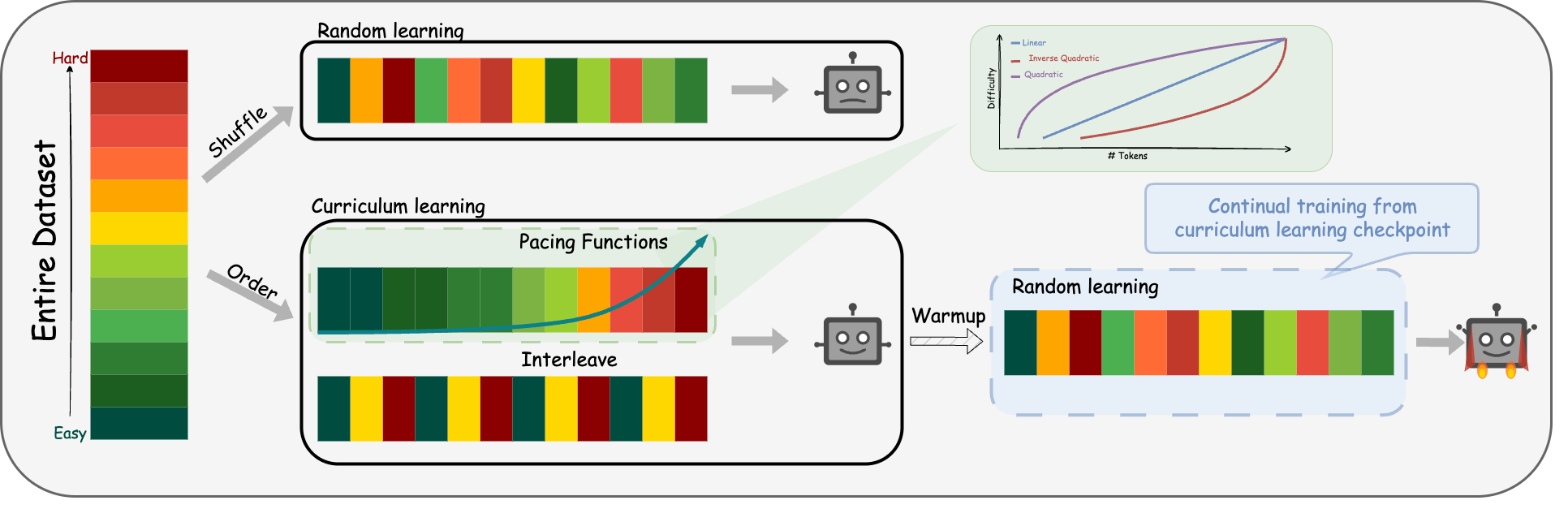}
  \caption{Curriculum learning for LLM pretraining organizes data from easy to hard based on a difficulty metric. This can involve strict ordering or progression governed by pacing functions such as linear, quadratic, or inverse quadratic. Alternatively, interleaved sampling strategies mix difficulty levels within each training segment. These strategies aim to enhance data efficiency and convergence. When using curriculum learning as warm-up followed by random training can further improve performance.}
  \label{fig:pacingfunc}
\end{figure*} 

To address this question, we conduct the first systematic investigation of curriculum learning for LLM pretraining. We evaluate three curriculum strategies \textbf{(i) vanilla curriculum learning} of a fixed dataset \cite{10.1145/1553374.1553380}, \textbf{(ii) sampling guided by pacing functions} \cite{wu2020curricula, nagatsuka2023length}, and \textbf{(iii) interleaved curricula} \cite{yang2024evaluating} that mix difficulty levels during training. We select 6 difficulty metrics out of 15 candidates through correlation analysis to characterize training data from multiple linguistic and information-theoretic perspectives. Using the CulturaX dataset \cite{nguyen-etal-2024-culturax}, we train a set of 0.5B - 3B parameter LMs on up to 100B tokens under three representative training scenarios: \textbf{limited data training}, \textbf{unlimited data training}, and \textbf{continual training}. We evaluate them on eight established benchmarks, spanning commonsense reasoning, language understanding, and reading comprehension.

Our findings demonstrate that curriculum learning can provide consistent advantages over random baselines under certain combinations of difficulty metrics and curriculum strategies, particularly in the early and mid phases of training. Models trained with curricula reached baseline performance 18–45\% faster, and we show that curriculum-based warmup can yield lasting performance gains even when followed by randomly sampled training with up to $3.5\%$ improvement. These results underscore the potential of curriculum design to enhance pretraining efficiency and open new directions for scalable, data-aware model development.

Our contributions are threefold: (1) a comprehensive study of curriculum learning in LLM pretraining under three realistic training scenarios with five data ordering strategies; (2) an empirical analysis of six difficulty metrics and their effects on model convergence and performance; and (3) evidence that curriculum-based warmup can serve as a practical mechanism for efficient model training. Our work provides actionable insights for improving the efficiency of LLM pretraining in both academic and industrial settings.

\section{Related Work}

\paragraph{Data-Efficient LLM Training}

Recent work on data-efficient LLM pretraining emphasizes pruning, reweighting, and selection to reduce training costs without sacrificing performance. Perplexity-based filtering \cite{marion2023less}, robust domain mixing (DoReMi) \cite{xie2023doremi}, and embedding or influence-based sampling \cite{tirumala2023d4, yu2024mates} all yield strong efficiency gains. Model-driven approaches like ASK-LLM and DENSITY leverage quality and diversity signals to outperform full-data baselines with fewer tokens \cite{sachdeva2024train}. Unlike these methods, which rely on sampling or reweighting data, we explore curriculum-based ordering over a pre-filtered dataset, making our approach complementary and orthogonal to prior works, providing a new dimension for improving training efficiency.

\paragraph{Curriculum Learning} 

Curriculum learning (CL), introduced by \cite{bengio2009curriculum}, improves convergence by training on increasingly difficult data. Early NLP applications include grammar induction and machine translation \cite{spitkovsky2009baby, zhang2018empirical}. Later work extended CL to LSTMs \cite{cirik2016visualizing} and transformers \cite{nagatsuka2023length}, though often limited to masked language modeling. Recent studies apply CL during LLM fine-tuning \cite{yang2024evaluating}, or explore data-efficient schedules via skill learning and model preference \cite{chen2023skill, zhang2025preference}. Other strategies vary input lengths or attention to reduce compute \cite{pouransari2024dataset, kim2024strategic}. However, prior work largely focuses on fine-tuning or narrow CL setups. Our work is the first to systematically study CL during LLM pretraining, evaluating multiple paradigms and difficulty metrics at scale.


\paragraph{Text Difficulty Estimation} 

Numerous studies explore text difficulty and quality in various applications. Length-based heuristics and term frequency are common proxies for complexity \cite{nagatsuka2023length, spitkovsky2009baby, liu2018curriculum}. Lexical diversity metrics like MTLD and vocd-D provide finer-grained signals, with MTLD noted for its robustness across corpora \cite{mccarthy2010mtld}. Information-theoretic measures such as compression ratio and entropy capture redundancy and quality \cite{yin2024entropy}, while recent work introduces perplexity-based preference modeling and diversity coefficients to assess conceptual variability \cite{zhang2025preference, miranda2023beyond}. Together, these works provide valuable candidates for exploring difficulty metrics in curriculum learning.  

\section{Methodology}
In this section, we discuss our choice of difficulty metrics, training scenarios, as well as the curriculum learning settings for our experiments.

\subsection{Difficulty Metrics}
To apply curriculum learning to LLM pretraining, we first constructed a diverse pool of 15 metrics, categorized into six conceptual dimensions: information density, lexical diversity, readability, fertility, model-perceived difficulty, and sequence length. We aimed to cover a broad range of candidate metrics while excluding task-specific and model-defined ones that are not universally adaptive or are computationally expensive. To ensure orthogonality and avoid redundancy, we performed a Spearman correlation analysis (see Appendix~\ref{app:correlation}) across these candidates. Unlike previous works, we prioritize metrics that are broadly representative of distinct linguistic and information-theoretic properties, easy to compute, independent of the choice of model and language, and incur minimal computational overhead during training (see Appendix~\ref{sec:overhead}).


From this analysis, we selected six representative and minimally correlated metrics that capture distinct aspects of textual complexity: \textbf{Compression Ratio} \cite{yin2024entropy}, measuring how compactly information is encoded, serving as a proxy for redundancy and structural regularity; \textbf{Fertility} \cite{ali2024tokenizer}, Measures tokenization complexity, defined as the average number of subword tokens per word; \textbf{Flesch Reading Ease} \cite{kincaid1975derivation}, capturing readability by estimating text comprehensibility; Measure of Textual Lexical Diversity \textbf{(MTLD)} \cite{mccarthy2010mtld}, captures lexical richness, offering a length-insensitive and robust estimate of vocabulary diversity; \textbf{Number of Tokens}, token level sequence length; \textbf{Perplexity}, a model-centric metric reflecting linguistic uncertainty as perceived by a pretrained language model. 

\subsection{Vanilla Curriculum Learning}

We define our first training scenario of limited data training:
\begin{hyp}[S\ref{hyp:first}]\label{hyp:first} We assume to have access to a limited, fixed set of pretraining data, all of which must be utilized during training.\end{hyp}

Under S\ref{hyp:first}, we adopt vanilla CL: a strict ordering of training samples from easy to hard \cite{10.1145/1553374.1553380}. To do this, we sort the data by its difficulty score in ascending order. The model is then trained on this ordered set. This setting introduces more difficult samples at every step, aiming to gradually increase the model's capacity to handle complexity by building on previously acquired knowledge.


\subsection{Curriculum Learning with Pacing}

To introduce pacing functions, we define our second training scenario of training from a large data pool:
\begin{hyp}[S\ref{hyp:second}]\label{hyp:second} We assume to have access to a large (effectively unlimited) dataset, from which we are allowed to sample training data up to a fixed training budget.\end{hyp}

Under S\ref{hyp:second}, instead of random sample training data, we sample data progressively using pacing functions that control the difficulty distribution over time. 

Following works like \cite{hacohen2019power, nagatsuka2023length}, we split the dataset into $N$ equally sized difficulty groups. At each training stage, the pacing function determines the number of samples drawn from each group. Sampling is performed randomly within groups to preserve diversity, while the pacing schedule governs the gradual transition from easy to hard examples.

We explore three types of pacing functions, as shown in Figure \ref{fig:pacingfunc}. Assume we have $N$ difficulty groups, let $T$ denote the total number of training tokens and $t_i$ the number of tokens allocated to group $i$.

\paragraph{Linear Pacing.} 
\begin{equation}
t_i = \frac{T}{N}
\end{equation}
The linear pacing function increases the difficulty level at a constant rate over time by an equal token allocation across groups.

\paragraph{Quadratic Pacing.} 
\begin{equation}
t_i = T \cdot \frac{(i + 1)^2}{\sum_{j=1}^{N} (j + 1)^2}
\end{equation}
The quadratic pacing function allocates a larger portion of the token budget to higher-difficulty groups in later stages. Compared to linear pacing, this results in a faster increase in difficulty over training. 


\paragraph{Inverse Quadratic Pacing.}
\begin{equation}
t_i = T \cdot \frac{(N - i)^2}{\sum_{j=0}^{N-1} (N - j)^2}
\end{equation}

Opposite to the quadratic approach, the inverse quadratic pacing function allocates more tokens to easier groups early on.

\subsubsection{{Interleaved Curriculum Learning}}
As a special case of pacing functions, we applied interleaved sampling. Rather than segmenting training into distinct difficulty phases, interleaved curricula maintain a continuous mix of difficulty levels throughout training as shown in Figure \ref{fig:pacingfunc}. The dataset is divided into $N$ equally sized difficulty groups, and training is organized into $I$ interleaves, each covering a token budget of $T/I$, where $T$ is the total training budget. Within each interleave, we apply a linear pacing function to sample data in an easy-to-hard order, so for the current group $i$ in interleave $j$, the number of tokens sampled $t_{i,j}$ is:

\begin{equation}
t_{i,j} = \frac{T}{I \cdot N}
\end{equation}

This setup allows the model to repeatedly encounter the full difficulty spectrum while preserving a progressive structure within each interleave. Such exposure aims to improve generalization by preventing overfitting to a narrow band. 



\begin{figure*}[t]
  \centering
  \includegraphics[width=\textwidth]{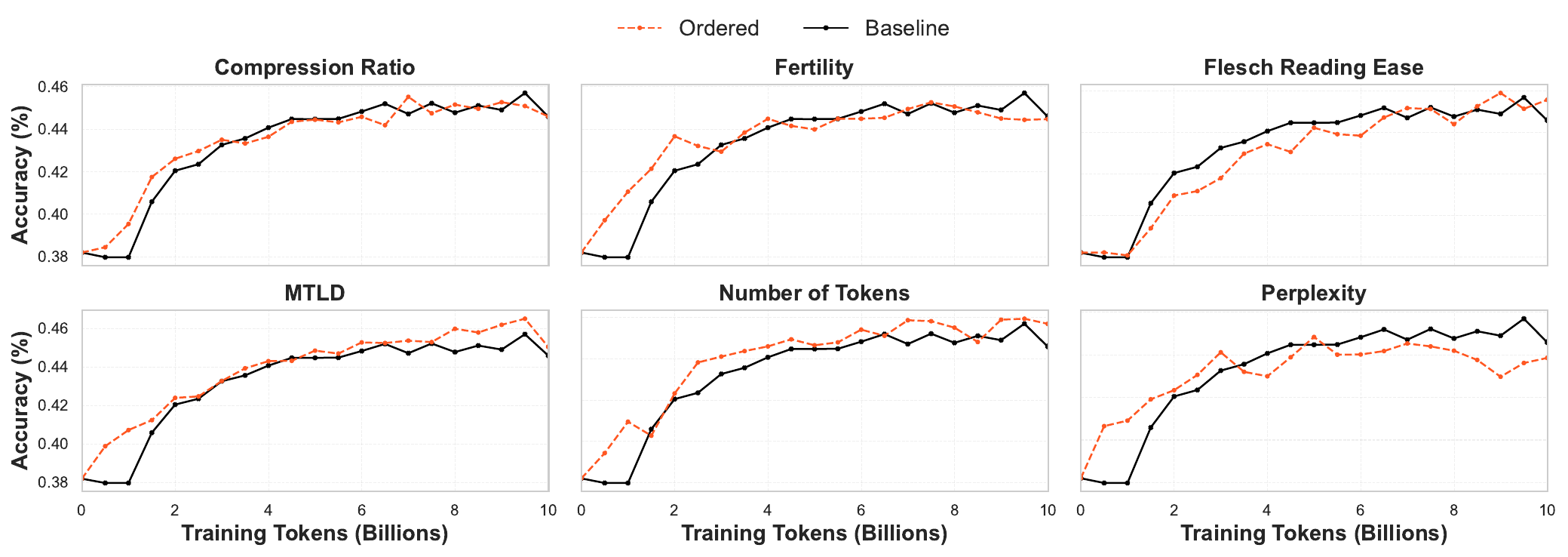}
  \caption{Evaluation accuracy of models trained with vanilla curriculum learning across 6 difficulty metrics, compared to a randomly shuffled baseline. CL settings offer a notable performance boost in early to mid stages, particularly under MTLD and Number of Tokens metrics, the models consistently outperform the baseline.}
  \label{fig:strict_order}
\end{figure*}

\subsection{Experimental setting}

\paragraph{Dataset} We use the English subset of CulturaX \cite{nguyen-etal-2024-culturax} for pretraining, selected for its quality, transparency, and robust preprocessing. CulturaX is created by extensive cleaning: language ID, URL filtering, metric-based heuristics, document refinement, and fuzzy deduplication via MinHash. These steps are guided by large-scale data inspection and diverse quality metrics. We use the dataset as released, without further preprocessing.


\paragraph{Model} We adopt a 0.5 billion parameter decoder-only language model for the main experiments based on the LLaMA3.2 \cite{grattafiori2024llama} architecture. The model comprises 16 transformer decoder layers with a hidden size of 1536 and 16 attention heads, using Multi-Query Attention (MQA) with 8 key-value groups. We apply rotary positional embeddings with a scaling factor of 1.0, $\theta = 10^5$, and a rotary dimension of 96. The feedforward network uses a SwiGLU activation and an intermediate size of 4096, and a maximum sequence length of 2048. For the 1B and 3B models, we use the same configuration as LLaMA3.2.


\paragraph{Training} 
We train all models using the AdamW optimizer \cite{loshchilov2017decoupled} with $\beta_1 = 0.9$, $\beta_2 = 0.95$, weight decay of 0.1, and a fixed learning rate of $2 \times 10^{-4}$. We apply gradient clipping with a max norm of 1.0, mixed-precision training (cbfloat16), and dynamic loss scaling. The global batch size is 1,081,344 tokens. Training is performed on the Condor Galaxy 2 AI supercomputer (see Appendix~\ref{sec:cerebras}), and evaluations are run on NVIDIA A10 GPUs. 

To ensure reproducibility across hardware, we conducted additional experiments comparing models trained on both platforms and observed consistent performance (Appendix~\ref{sec:cerebras}).



\noindent\textbf{Evaluation} We evaluate each model using LM-Eval \cite{eval-harness}, on a comprehensive set of 8 benchmarks from different categories, we report the average of all benchmarks. The details of the benchmarks are listed below:
\begin{itemize}
    \item \textbf{Commonsense Reasoning}: PIQA \cite{bisk2020piqa}, COPA \cite{gordon-etal-2012-semeval}, OpenBookQA \cite{mihaylov2018can}.
    \item \textbf{Language Understanding}: Hellaswag \cite{zellers2019hellaswag}, WinoGrande \cite{sakaguchi2021winogrande}, xwinograd\_en \cite{tikhonov2021heads}.
    \item \textbf{Reading Comprehension}: BoolQ \cite{clark2019boolq}.
    \item \textbf{World Knowledge}: ArcChallenge \cite{clark2018think}.
\end{itemize}

All benchmarks are evaluated with 0-shot.

\begin{figure*}[t]
  \centering
  \includegraphics[width=\linewidth]{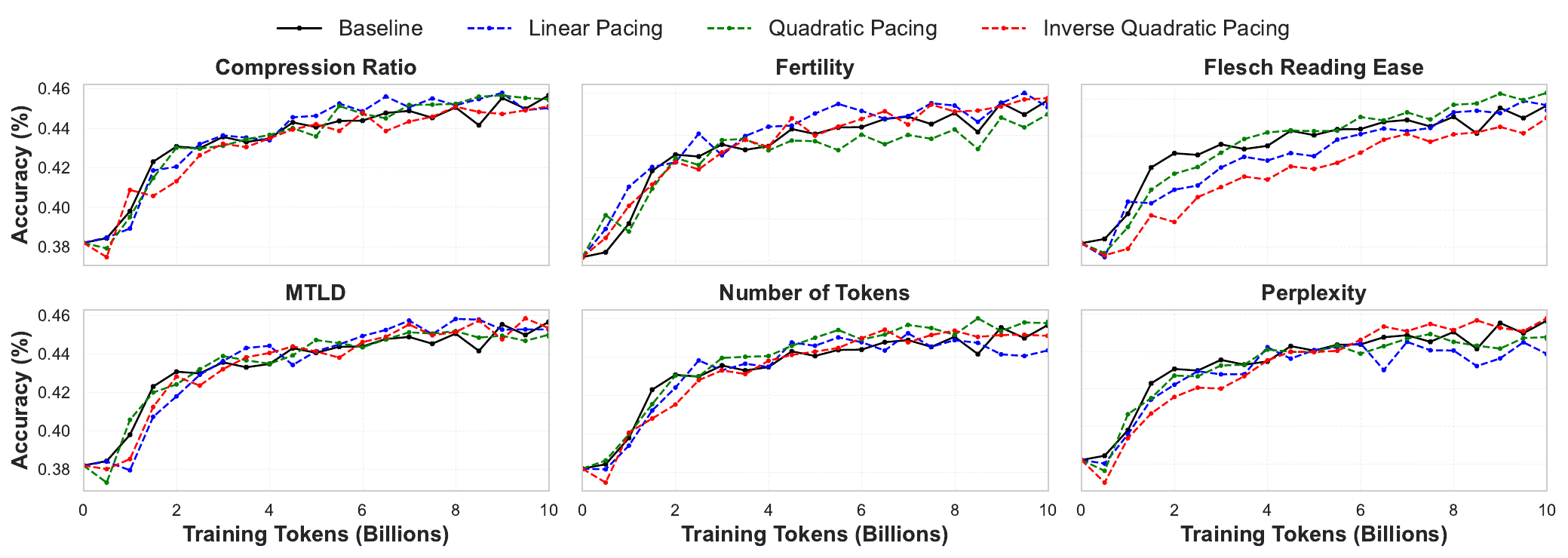}
  \caption{Evaluation accuracy of models trained with pacing-based curriculum learning across 3 pacing functions and 6 difficulty metrics. Compression Ratio, MTLD, and Fertility benefit most from linear pacing. Flesch Reading Ease and Number of Tokens yield best performance with quadratic pacing, with Flesch Reading Ease showing a consistent upward trend across all pacing functions.}
  \label{fig:pacing}
\end{figure*}

\section{Experiments}
In this section, we show our experiments in detail on the curriculum settings and report the average evaluation accuracy, 
since training or validation loss reflects token prediction accuracy on current dataset but poorly captures generalization, often showing weak correlation with downstream task performance \cite{liu2023same, hu2024can}. In contrast, benchmark evaluations directly assess real-world capabilities like reasoning and comprehension, offering a more reliable basis for comparing LLMs \cite{tay2021scale}.

\subsection{Experiment 1: Strict ordering}
\label{exp1}
\paragraph{Setup} Under S\ref{hyp:first}, we first construct a fixed training set. Following the scaling laws of model performance~\cite{kaplan2020scaling}, we sample 10B tokens from CulturaX to pretrain a 0.5B parameter model. For the 6 selected difficulty metrics, we construct one set for each by ordering the data based on the difficulty score. 
As the baseline, we randomly shuffled this fixed dataset. We pretrain one model for each of the 7 training sets.

\begin{figure*}[t]
  \centering
  \includegraphics[width=\textwidth]{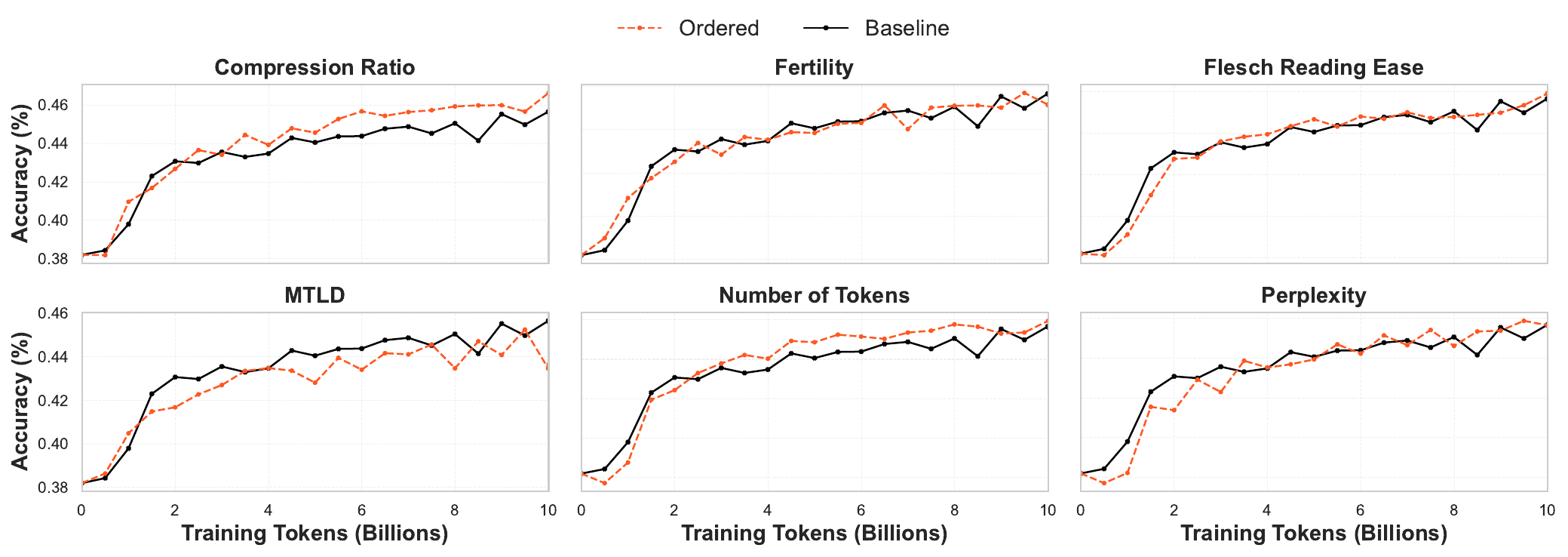}
  \caption{Evaluation accuracy of models trained with interleaved curriculum learning across 6 difficulty metrics. Interleaving consistently boosts performance for Compression Ratio and Number of Tokens, with the rest similar or slightly worse than the baseline.}
  \label{fig:interleave}
\end{figure*}

\paragraph{Results}
\textit{At the early stage of training, easier samples—according to most difficulty metrics—facilitate more efficient learning.}

Figure \ref{fig:strict_order} presents the average accuracy of all benchmarks for models trained with strict ordering under 6 difficulty metrics, compared to the baseline. Except for Flesch Reading Ease, models trained on CL subsets consistently outperform the random baseline during the early stages of training (up to approximately 4B tokens). The advantage of curriculum learning narrows as training progresses, with performances converging or remaining slightly higher than the baseline by the end of training.

Among the 6 metrics, MTLD and Number of Tokens settings maintain benefits throughout the training, with MTLD achieving $1.8\%$ higher best performance than baseline using $17.9\%$ fewer training steps, and Number of Tokens achieves similar performance but with $27.5\%$ fewer steps.

Interestingly, the Flesch Reading Ease setting initially underperforms the baseline but gradually catches up at a constant rate in later stages, showing a similar pattern as we observed in the following experiments, where we will provide more analysis. 

Perplexity-based ordering, although showing strong early gains, exhibits a noticeable drop in performance during the later phases. We hypothesize that high-perplexity samples that are often both difficult and noisy, which could degrade the model’s learning stability. 



\subsection{Experiment 2: Pacing Functions}

\paragraph{Setup}In this setting, we first follow S\ref{hyp:second} to randomly sample and shuffle 10B tokens as the baseline. Then for each difficulty metric, we first partition the dataset into 10 difficulty groups of equal size. We then apply linear, quadratic, and inverse quadratic pacing functions to sample data from each group, gradually building the training set to 10B tokens. We train one model for each combination of pacing function and difficulty metric.

\paragraph{Results}\textit{Compared with vanilla curriculum learning, a steady, progressive exposure to increasingly difficult
groups while keeping variation within groups improves model performance more effectively for metrics capturing linguistic richness. And model trained on data ordered by Flesch Reading Ease show consistent improvement without a sign of convergence.}

Figure~\ref{fig:pacing} shows results grouped by difficulty metrics. CL settings differs more to the baseline in early to mid training stages, which aligns with the findings in section \ref{exp1}.

Specifically, Compression Ratio, Fertility, and MTLD—metrics tied to linguistic richness and redundancy—benefit most from linear pacing, especially in the mid-training stage. Compression Ratio reaches baseline peak accuracy with $39.5\%$ fewer steps; MTLD with $31.7\%$ fewer steps; and Fertility achieves $99.5\%$ of best baseline performance using $45.3\%$ fewer steps.

For the Number of Tokens setting, all pacing strategies outperform the baseline between 3B–8B tokens, with quadratic pacing delivering the most stable gains: it achieves the baseline's performance in $29.8\%$ fewer steps, reaches $99.5\%$ of the peak in $44.8\%$ fewer steps, and ultimately surpasses the baseline with a final score that is $1.1\%$ higher.

For Flesch Reading Ease, interestingly, 3 pacing functions show a similar trend as shown in Experiment 1: model performance increases with constant speed and doesn't show a sign of convergence, with quadratic pacing yielding the best result: $1.6\%$ above the baseline at the final checkpoint.

In contrast, Perplexity performs best under inverse quadratic pacing, while linear and quadratic pacing fall behind, likely due to noisy high-perplexity samples—consistent with findings from Experiment 1.

\subsection{Experiment 3: Interleave Curriculum Learning}
\paragraph{Setup}In this setting, we apply interleaved curriculum learning under S\ref{hyp:second}. As before, for each difficulty metric, the dataset is partitioned into 10 equal-sized difficulty groups. We then construct 10 interleaved subsets, each consisting of 1B tokens sampled linearly across all difficulty groups. These subsets are concatenated to form a 10B-token training set, allowing the model to encounter a mix of difficulties throughout training while preserving a progressive structure within each interleave.

\paragraph{Results}\textit{Only the Compression Ratio and Number of Tokens settings showed a consistent advantage over the baseline, while others remain similar or have slightly worse performance}

Figure~\ref{fig:interleave} reports the averaged accuracy. Our results show that the compression ratio setting benefits the most from interleaved CL, where the model outperforms the baseline throughout the training by a large margin, with $2.2\%$ higher performance achieved, and the best random baseline performance is reached with $41.7\%$ fewer steps. A similar pattern can be seen for the Number of Tokens setting, where the model consistently outperforms the baseline and reached the best baseline performance with $23.5\%$ less data. 

In contrast, other metrics yield nearly the same performance as the random baseline and the MTLD setting has worse performance during training.

\subsection{Curriculum Learning as model warmup}
Inspired by the fact that the curriculum learning generally speeds up the early to mid training stage and training with CL is usually one-shot: once the training set is organized by CL, it is difficult to add new data to the dataset unless we reorganize the enlarged dataset again. So we consider the third training scenario of continual training:

\begin{hyp}[S\ref{hyp:third}]\label{hyp:third} The training consists of multi phases, where in the first phase the model is trained with a data budget, for later phases it can be further trained on more data.\end{hyp}

The S\ref{hyp:third} involves cases like continual training where CL can only apply to the first phase because organizing data from easy to hard is one-shot, and massive data training where organizing the entire dataset with CL is computationally hard.

\paragraph{Setup} In these cases, we consider CL as a warmup phase for model training where we first train the model with a fixed data organized under CL, then continue training it with more data under the conventional training with shuffling.

For the CL training phase, we choose 3 CL settings that previously showed promising results: Compression Ratio, Number of Tokens with vanilla CL and Flesch Reading Ease with quadratic pacing. For the second training phase, we continue training the models of each setting as well as the baseline using the same 10B new data which is randomly shuffled. For curriculum learning settings, we choose two checkpoints to perform continual training: the best one and the last one, if they are not the same.

\paragraph{Results} \textit{Using CL as warmup consistently improves performance in later training phases by a large margin, allowing the model reach a higher convergence point.}

\begin{figure}[t]
  \centering
  \includegraphics[width=\columnwidth]{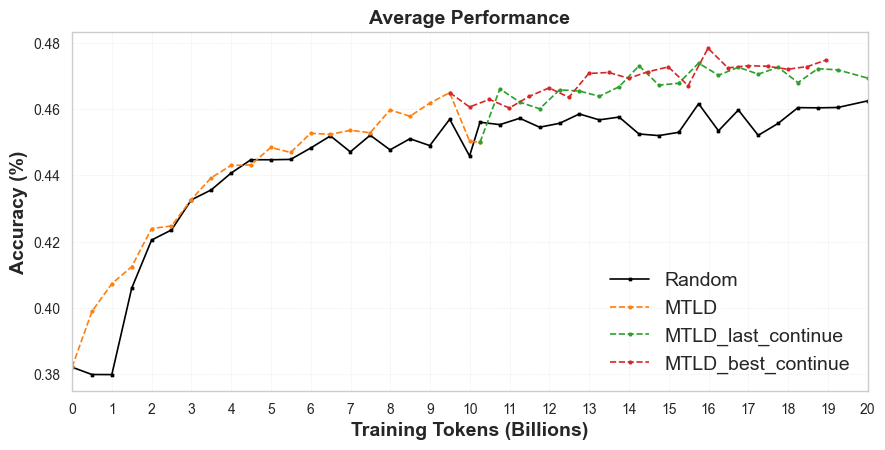}
  \caption{Performance of models trained with curriculum learning as a warmup phase, followed by continued training on randomly shuffled data, using MTLD as the difficulty metric, both best and last checkpoint warmups yield sustained advantages over the baseline.}
  \label{fig:mtld_continue}
\end{figure}

We show the result for the MTLD setting in Figure \ref{fig:mtld_continue}: models with a transition from CL to random training maintain a consistent performance advantage over the baseline. While the baseline performance converges, the warmup setting keeps improving regardless of starting from the best or last checkpoint. Specifically, the starting from the best and last lead to $3.5\%$ and $2.6\%$ higher than the baseline, respectively, and the baseline never surpasses the first phase of CL training with double the data. More results are in Appendix \ref{sec:warmup}.

\section{Ablation Study}


\subsection{Number of difficulty groups}
In our main experiments, we chose 10 difficulty groups when splitting the dataset to balance smooth difficulty transitions with sufficient intra-group diversity for effective learning. 

Following \cite{nagatsuka2023length}, we split the data into 3 difficulty groups as easy, middle, and hard, to study the impact of fewer groups on the linear pacing case. Detailed results are shown in Appendix \ref{sec:numgroups}, where using fewer groups leads to nearly equal or worse model performance.

We also tested the case with 20 groups in the early stage of training, showing no improvement compared to 10 groups. This indicates that when applying pacing functions, the results are less sensitive to the choice of the number of groups. We assume fewer groups lead to coarser transitions and potential overfitting, while more groups exhibit diminishing returns, behaving similarly to strict ordering. Thus, we use 10 groups in our experiments, which act as a balance between smooth difficulty transitions and sufficient intra-group diversity.

\subsection{Robustness of Curriculum Learning}

To evaluate the robustness of our curriculum learning strategies, we repeat the baseline and key experiments across six high-performing settings—spanning different difficulty metrics and pacing types—twice, independently for all stages from data sampling to model training. A detailed analysis is provided in the Appendix \ref{sec:appendix2}.

We observe consistent trends across runs, \textit{the previous findings with the CL settings still hold on average, even when accounting for variability across runs.} This low variance across repetitions confirms that the observed gains are stable and not due to chance, demonstrating that curriculum learning is a reliable approach for improving LLM pretraining efficiency.

\subsection{Data sampling vs Curriculum Learning}

We observed that in some settings, the model converges faster than the baseline but shows a performance drop at the last steps. The best practice is to select the best checkpoint.  
However, in CL, later-stage data can sometimes act as noisy or overly difficult samples, which may not further improve the model. So we ask the following question: \textit{are the early convergence or performance peaks results from the data filtering effect of CL?} 

To answer the question, we conduct additional experiments on selected settings that exhibit early convergence: MTLD, Number of Tokens with vanilla CL and Compression Ratio with linear pacing. Specifically, we extract training data used before the best-performing checkpoint, randomly shuffle it, and train the model again on this shuffled subset. A detailed comparison of the model’s performance on the shuffled subset versus the original ordered setting is provided in the Appendix \ref{sec:samplevsorder}.

Our results show that using \textit{the early stage data of curriculum learning indeed act as a data filtering and lead to improvement compared with the random baseline, but still fall behind the ordered data setting}, showing the effectiveness of data ordering during the training of the model.

\subsection{Scale Curriculum Learning}
In order to investigate how well our findings scale up with model size, we train a set of the 1B and 3B models with 20B and 40B tokens respectively on: (i) the vanilla CL setting, (ii) Flesch Reading Ease with quadratic pacing and (iii) CL as warmup from MTLD vanilla CL and Flesch Reading Ease with quadratic pacing setting. Full training configurations and results are presented in Appendix~\ref{sec:scale}.

Our results indicate that \textit{the benefits of curriculum learning extend to larger models.} In particular, the warmup setting consistently delivers sustained performance improvements throughout training, demonstrating the generalizability of curriculum learning across model scales and training scenarios

\subsection{Scale Curriculum Learning}
In order to investigate how well our findings scale up with model size and dataset size, we train a set of the 1B and 3B models with 20B and 40B tokens respectively on: (i) the vanilla CL setting, (ii) Flesch Reading Ease with quadratic pacing and (iii) CL as warmup from MTLD vanilla CL and Flesch Reading Ease with quadratic pacing setting. In addition, we extend the CL as warmup setting to 100B training tokens, following prior work suggesting the emergence of new capabilities at this scale. Full training configurations and results are presented in Appendix~\ref{sec:scale}.

Our results indicate that \textit{the benefits of curriculum learning extend to larger models and larger datasets.} In particular, the warmup setting consistently delivers sustained performance improvements throughout training, outperforming the baseline across all checkpoints. When trained to 100B tokens, curriculum learning as warmup achieves an improvement of approximately 3.1\%, demonstrating the generalizability and robustness of curriculum learning across model scales, dataset sizes, and training scenarios.

\section{Discussion and Conclusion}

Our work presents the first systematic exploration of curriculum learning for large language model pretraining, comparing vanilla, pacing-based, and interleaved curricula across six difficulty metrics. We demonstrate that \textbf{CL consistently accelerates convergence—reducing training steps by 18–45\%—and improves final performance by up to 3.5\% when applied as a warmup before random sampling.} Notably, this warmup strategy emerges as the most robust and practical setting, showing that CL not only speeds early and mid-stage training but also yields lasting gains in later phases.


Among the examined metrics, \textbf{Compression Ratio, MTLD, and Flesch Reading Ease stand out as the most effective indicators, linking linguistic diversity and information density to better learning progress.} In contrast, perplexity proves to be less effective: high-perplexity samples are placed at the end of training and are often noisy and low quality, which can hinder late-stage optimization. Linear pacing and interleaved strategies further enhance generalization by gradually and repeatedly exposing the model to increasing difficulty levels.

Importantly, \textbf{our approach introduces no computational overhead}—difficulty metrics are lightweight, model- and language-independent, and can be precomputed efficiently. CL operates purely on data ordering, orthogonal to data selection or pruning techniques, and thus integrates seamlessly into existing pretraining pipelines.

Overall, curriculum learning offers a simple yet powerful mechanism to improve training efficiency at scale. It can achieve both faster training and improved performance, even be used as a warmup phase, paving the way toward more data- and compute-efficient LLM development. Future work should explore adaptive and model-aware curricula for even greater efficiency gains.

\section*{Limitations}
This work focuses on LLM pretraining via curriculum learning by training 0.5B and 1B parameter models on various combinations of curriculum strategies and difficulty metrics. While our findings consistently demonstrate the benefits of curriculum learning (CL) across training scenarios, several limitations remain:

First, our experiments are constrained to decoder-only architectures (LLaMA3.2-like models) and English-language data (CulturaX subset), which may limit the generalizability of our results to encoder-based models or multilingual settings. Future work should evaluate CL across diverse model families and linguistic contexts.

Second, we apply static, precomputed difficulty scores and do not explore adaptive or dynamic curricula that respond to the model’s evolving capabilities. More sophisticated scheduling techniques, such as model-aware or task-informed pacing, could yield further gains.

Third, while we test six difficulty metrics, our metric selection is based on correlation analysis, and some potentially useful signals (e.g., syntactic depth or coherence) were not explored. Additionally, metrics like perplexity may conflate difficulty with noise, reducing their reliability.

Fourth, this work focuses exclusively on the pretraining phase and does not examine the impact of curriculum learning on downstream fine-tuning. Evaluating how curriculum-pretrained models transfer to various fine-tuning regimes remains an important direction for future study.

\bibliography{googlescholar,aclanthology}

@article{soviany2022curriculum,
  title={Curriculum learning: A survey},
  author={Soviany, Petru and Ionescu, Radu Tudor and Rota, Paolo and Sebe, Nicu},
  journal={International Journal of Computer Vision},
  volume={130},
  number={6},
  pages={1526--1565},
  year={2022},
  publisher={Springer}
}

@inproceedings{10.1145/1553374.1553380,
author = {Bengio, Yoshua and Louradour, J\'{e}r\^{o}me and Collobert, Ronan and Weston, Jason},
title = {Curriculum learning},
year = {2009},
isbn = {9781605585161},
publisher = {Association for Computing Machinery},
address = {New York, NY, USA},
url = {https://doi.org/10.1145/1553374.1553380},
doi = {10.1145/1553374.1553380},
booktitle = {Proceedings of the 26th Annual International Conference on Machine Learning},
pages = {41–48},
numpages = {8},
location = {Montreal, Quebec, Canada},
series = {ICML '09}
}

@article{loshchilov2017decoupled,
  title={Decoupled weight decay regularization},
  author={Loshchilov, Ilya and Hutter, Frank},
  journal={arXiv preprint arXiv:1711.05101},
  year={2017}
}

@article{kaplan2020scaling,
  title={Scaling laws for neural language models},
  author={Kaplan, Jared and McCandlish, Sam and Henighan, Tom and Brown, Tom B and Chess, Benjamin and Child, Rewon and Gray, Scott and Radford, Alec and Wu, Jeffrey and Amodei, Dario},
  journal={arXiv preprint arXiv:2001.08361},
  year={2020}
}

@misc{eval-harness,
  author       = {Gao, Leo and Tow, Jonathan and Abbasi, Baber and Biderman, Stella and Black, Sid and DiPofi, Anthony and Foster, Charles and Golding, Laurence and Hsu, Jeffrey and Le Noac'h, Alain and Li, Haonan and McDonell, Kyle and Muennighoff, Niklas and Ociepa, Chris and Phang, Jason and Reynolds, Laria and Schoelkopf, Hailey and Skowron, Aviya and Sutawika, Lintang and Tang, Eric and Thite, Anish and Wang, Ben and Wang, Kevin and Zou, Andy},
  title        = {The Language Model Evaluation Harness},
  month        = 07,
  year         = 2024,
  publisher    = {Zenodo},
  version      = {v0.4.3},
  doi          = {10.5281/zenodo.12608602},
  url          = {https://zenodo.org/records/12608602}
}

@inproceedings{bisk2020piqa,
  title={Piqa: Reasoning about physical commonsense in natural language},
  author={Bisk, Yonatan and Zellers, Rowan and Gao, Jianfeng and Choi, Yejin and others},
  booktitle={Proceedings of the AAAI conference on artificial intelligence},
  volume={34},
  pages={7432--7439},
  year={2020}
}

@article{mihaylov2018can,
  title={Can a suit of armor conduct electricity? a new dataset for open book question answering},
  author={Mihaylov, Todor and Clark, Peter and Khot, Tushar and Sabharwal, Ashish},
  journal={arXiv preprint arXiv:1809.02789},
  year={2018}
}

@article{zellers2019hellaswag,
  title={Hellaswag: Can a machine really finish your sentence?},
  author={Zellers, Rowan and Holtzman, Ari and Bisk, Yonatan and Farhadi, Ali and Choi, Yejin},
  journal={arXiv preprint arXiv:1905.07830},
  year={2019}
}

@article{sakaguchi2021winogrande,
  title={Winogrande: An adversarial winograd schema challenge at scale},
  author={Sakaguchi, Keisuke and Bras, Ronan Le and Bhagavatula, Chandra and Choi, Yejin},
  journal={Communications of the ACM},
  volume={64},
  number={9},
  pages={99--106},
  year={2021},
  publisher={ACM New York, NY, USA}
}

@misc{tikhonov2021heads,
    title={It's All in the Heads: Using Attention Heads as a Baseline for Cross-Lingual Transfer in Commonsense Reasoning},
    author={Alexey Tikhonov and Max Ryabinin},
    year={2021},
    eprint={2106.12066},
    archivePrefix={arXiv},
    primaryClass={cs.CL}
}

@article{clark2019boolq,
  title={Boolq: Exploring the surprising difficulty of natural yes/no questions},
  author={Clark, Christopher and Lee, Kenton and Chang, Ming-Wei and Kwiatkowski, Tom and Collins, Michael and Toutanova, Kristina},
  journal={arXiv preprint arXiv:1905.10044},
  year={2019}
}

@article{clark2018think,
  title={Think you have solved question answering? try arc, the ai2 reasoning challenge},
  author={Clark, Peter and Cowhey, Isaac and Etzioni, Oren and Khot, Tushar and Sabharwal, Ashish and Schoenick, Carissa and Tafjord, Oyvind},
  journal={arXiv preprint arXiv:1803.05457},
  year={2018}
}

@article{marion2023less,
  title={When less is more: Investigating data pruning for pretraining llms at scale},
  author={Marion, Max and {\"U}st{\"u}n, Ahmet and Pozzobon, Luiza and Wang, Alex and Fadaee, Marzieh and Hooker, Sara},
  journal={arXiv preprint arXiv:2309.04564},
  year={2023}
}

@article{xie2023doremi,
  title={Doremi: Optimizing data mixtures speeds up language model pretraining},
  author={Xie, Sang Michael and Pham, Hieu and Dong, Xuanyi and Du, Nan and Liu, Hanxiao and Lu, Yifeng and Liang, Percy S and Le, Quoc V and Ma, Tengyu and Yu, Adams Wei},
  journal={Advances in Neural Information Processing Systems},
  volume={36},
  pages={69798--69818},
  year={2023}
}

@article{tirumala2023d4,
  title={D4: Improving llm pretraining via document de-duplication and diversification},
  author={Tirumala, Kushal and Simig, Daniel and Aghajanyan, Armen and Morcos, Ari},
  journal={Advances in Neural Information Processing Systems},
  volume={36},
  pages={53983--53995},
  year={2023}
}

@article{sorscher2022beyond,
  title={Beyond neural scaling laws: beating power law scaling via data pruning},
  author={Sorscher, Ben and Geirhos, Robert and Shekhar, Shashank and Ganguli, Surya and Morcos, Ari},
  journal={Advances in Neural Information Processing Systems},
  volume={35},
  pages={19523--19536},
  year={2022}
}

@inproceedings{longpre2024pretrainer,
  title={A pretrainer’s guide to training data: Measuring the effects of data age, domain coverage, quality, \& toxicity},
  author={Longpre, Shayne and Yauney, Gregory and Reif, Emily and Lee, Katherine and Roberts, Adam and Zoph, Barret and Zhou, Denny and Wei, Jason and Robinson, Kevin and Mimno, David and others},
  booktitle={Proceedings of the 2024 Conference of the North American Chapter of the Association for Computational Linguistics: Human Language Technologies (Volume 1: Long Papers)},
  pages={3245--3276},
  year={2024}
}

@article{sachdeva2024train,
  title={How to train data-efficient llms},
  author={Sachdeva, Noveen and Coleman, Benjamin and Kang, Wang-Cheng and Ni, Jianmo and Hong, Lichan and Chi, Ed H and Caverlee, James and McAuley, Julian and Cheng, Derek Zhiyuan},
  journal={arXiv preprint arXiv:2402.09668},
  year={2024}
}

@article{yu2024mates,
  title={Mates: Model-aware data selection for efficient pretraining with data influence models},
  author={Yu, Zichun and Das, Spandan and Xiong, Chenyan},
  journal={Advances in Neural Information Processing Systems},
  volume={37},
  pages={108735--108759},
  year={2024}
}

@inproceedings{bengio2009curriculum,
  title={Curriculum learning},
  author={Bengio, Yoshua and Louradour, J{\'e}r{\^o}me and Collobert, Ronan and Weston, Jason},
  booktitle={Proceedings of the 26th annual international conference on machine learning},
  pages={41--48},
  year={2009}
}

@article{spitkovsky2009baby,
  title={Baby Steps: How “Less is More” in unsupervised dependency parsing},
  author={Spitkovsky, Valentin I and Alshawi, Hiyan and Jurafsky, Daniel},
  journal={NIPS: Grammar Induction, Representation of Language and Language Learning},
  pages={1--10},
  year={2009}
}

@article{cirik2016visualizing,
  title={Visualizing and understanding curriculum learning for long short-term memory networks},
  author={Cirik, Volkan and Hovy, Eduard and Morency, Louis-Philippe},
  journal={arXiv preprint arXiv:1611.06204},
  year={2016}
}

@article{zhang2018empirical,
  title={An empirical exploration of curriculum learning for neural machine translation},
  author={Zhang, Xuan and Kumar, Gaurav and Khayrallah, Huda and Murray, Kenton and Gwinnup, Jeremy and Martindale, Marianna J and McNamee, Paul and Duh, Kevin and Carpuat, Marine},
  journal={arXiv preprint arXiv:1811.00739},
  year={2018}
}

@inproceedings{liu2018curriculum,
  title={Curriculum Learning for Natural Answer Generation.},
  author={Liu, Cao and He, Shizhu and Liu, Kang and Zhao, Jun and others},
  booktitle={IJCAI},
  pages={4223--4229},
  year={2018}
}

@article{yang2024evaluating,
  title={Evaluating Fine-Tuning Efficiency of Human-Inspired Learning Strategies in Medical Question Answering},
  author={Yang, Yushi and Bean, Andrew M and McCraith, Robert and Mahdi, Adam},
  journal={arXiv preprint arXiv:2408.07888},
  year={2024}
}

@article{chen2023skill,
  title={Skill-it! a data-driven skills framework for understanding and training language models},
  author={Chen, Mayee and Roberts, Nicholas and Bhatia, Kush and Wang, Jue and Zhang, Ce and Sala, Frederic and R{\'e}, Christopher},
  journal={Advances in Neural Information Processing Systems},
  volume={36},
  pages={36000--36040},
  year={2023}
}

@article{pouransari2024dataset,
  title={Dataset decomposition: Faster llm training with variable sequence length curriculum},
  author={Pouransari, Hadi and Li, Chun-Liang and Chang, Jen-Hao and Anasosalu Vasu, Pavan Kumar and Koc, Cem and Shankar, Vaishaal and Tuzel, Oncel},
  journal={Advances in Neural Information Processing Systems},
  volume={37},
  pages={36121--36147},
  year={2024}
}

@article{kim2024strategic,
  title={Strategic Data Ordering: Enhancing Large Language Model Performance through Curriculum Learning},
  author={Kim, Jisu and Lee, Juhwan},
  journal={arXiv preprint arXiv:2405.07490},
  year={2024}
}

@article{mccarthy2010mtld,
  title={MTLD, vocd-D, and HD-D: A validation study of sophisticated approaches to lexical diversity assessment},
  author={McCarthy, Philip M and Jarvis, Scott},
  journal={Behavior research methods},
  volume={42},
  number={2},
  pages={381--392},
  year={2010},
  publisher={Springer}
}

@article{zhang2025preference,
  title={Preference Curriculum: LLMs Should Always Be Pretrained on Their Preferred Data},
  author={Zhang, Xuemiao and Xu, Liangyu and Duan, Feiyu and Zhou, Yongwei and Wang, Sirui and Weng, Rongxiang and Wang, Jingang and Cai, Xunliang},
  journal={arXiv preprint arXiv:2501.13126},
  year={2025}
}

@article{miranda2023beyond,
  title={Beyond Scale: The Diversity Coefficient as a Data Quality Metric for Variability in Natural Language Data},
  author={Miranda, Brando and Lee, Alycia and Sundar, Sudharsan and Casasola, Allison and Koyejo, Sanmi},
  journal={arXiv preprint arXiv:2306.13840},
  year={2023}
}

@article{kumar2010self,
  title={Self-paced learning for latent variable models},
  author={Kumar, M and Packer, Benjamin and Koller, Daphne},
  journal={Advances in neural information processing systems},
  volume={23},
  year={2010}
}

@article{sinha2020curriculum,
  title={Curriculum by smoothing},
  author={Sinha, Samarth and Garg, Animesh and Larochelle, Hugo},
  journal={Advances in Neural Information Processing Systems},
  volume={33},
  pages={21653--21664},
  year={2020}
}

@article{achiam2023gpt,
  title={Gpt-4 technical report},
  author={Achiam, Josh and Adler, Steven and Agarwal, Sandhini and Ahmad, Lama and Akkaya, Ilge and Aleman, Florencia Leoni and Almeida, Diogo and Altenschmidt, Janko and Altman, Sam and Anadkat, Shyamal and others},
  journal={arXiv preprint arXiv:2303.08774},
  year={2023}
}

@article{anil2023palm,
  title={Palm 2 technical report},
  author={Anil, Rohan and Dai, Andrew M and Firat, Orhan and Johnson, Melvin and Lepikhin, Dmitry and Passos, Alexandre and Shakeri, Siamak and Taropa, Emanuel and Bailey, Paige and Chen, Zhifeng and others},
  journal={arXiv preprint arXiv:2305.10403},
  year={2023}
}

@inproceedings{hacohen2019power,
  title={On the power of curriculum learning in training deep networks},
  author={Hacohen, Guy and Weinshall, Daphna},
  booktitle={International conference on machine learning},
  pages={2535--2544},
  year={2019},
  organization={PMLR}
}

@article{wu2020curricula,
  title={When do curricula work?},
  author={Wu, Xiaoxia and Dyer, Ethan and Neyshabur, Behnam},
  journal={arXiv preprint arXiv:2012.03107},
  year={2020}
}

@article{nagatsuka2023length,
  title={Length-based curriculum learning for efficient pre-training of language models},
  author={Nagatsuka, Koichi and Broni-Bediako, Clifford and Atsumi, Masayasu},
  journal={New Generation Computing},
  volume={41},
  number={1},
  pages={109--134},
  year={2023},
  publisher={Springer}
}

@article{yin2024entropy,
  title={Entropy law: The story behind data compression and llm performance},
  author={Yin, Mingjia and Wu, Chuhan and Wang, Yufei and Wang, Hao and Guo, Wei and Wang, Yasheng and Liu, Yong and Tang, Ruiming and Lian, Defu and Chen, Enhong},
  journal={arXiv preprint arXiv:2407.06645},
  year={2024}
}

@inproceedings{ali2024tokenizer,
  title={Tokenizer choice for llm training: Negligible or crucial?},
  author={Ali, Mehdi and Fromm, Michael and Thellmann, Klaudia and Rutmann, Richard and L{\"u}bbering, Max and Leveling, Johannes and Klug, Katrin and Ebert, Jan and Doll, Niclas and Buschhoff, Jasper and others},
  booktitle={Findings of the Association for Computational Linguistics: NAACL 2024},
  pages={3907--3924},
  year={2024}
}

@book{kincaid1975derivation,
  title={Derivation of New Readability Formulas: (automated Readability Index, Fog Count and Flesch Reading Ease Formula) for Navy Enlisted Personnel},
  author={Kincaid, J.P.},
  series={Research Branch report},
  url={https://books.google.fr/books?id=4tjroQEACAAJ},
  year={1975},
  publisher={Chief of Naval Technical Training, Naval Air Station Memphis}
}

@article{grattafiori2024llama,
  title={The llama 3 herd of models},
  author={Grattafiori, Aaron and Dubey, Abhimanyu and Jauhri, Abhinav and Pandey, Abhinav and Kadian, Abhishek and Al-Dahle, Ahmad and Letman, Aiesha and Mathur, Akhil and Schelten, Alan and Vaughan, Alex and others},
  journal={arXiv preprint arXiv:2407.21783},
  year={2024}
}

@inproceedings{liu2023same,
  title={Same pre-training loss, better downstream: Implicit bias matters for language models},
  author={Liu, Hong and Xie, Sang Michael and Li, Zhiyuan and Ma, Tengyu},
  booktitle={International Conference on Machine Learning},
  pages={22188--22214},
  year={2023},
  organization={PMLR}
}

@article{hu2024can,
  title={Can Perplexity Reflect Large Language Model's Ability in Long Text Understanding?},
  author={Hu, Yutong and Huang, Quzhe and Tao, Mingxu and Zhang, Chen and Feng, Yansong},
  journal={arXiv preprint arXiv:2405.06105},
  year={2024}
}

@article{tay2021scale,
  title={Scale efficiently: Insights from pre-training and fine-tuning transformers},
  author={Tay, Yi and Dehghani, Mostafa and Rao, Jinfeng and Fedus, William and Abnar, Samira and Chung, Hyung Won and Narang, Sharan and Yogatama, Dani and Vaswani, Ashish and Metzler, Donald},
  journal={arXiv preprint arXiv:2109.10686},
  year={2021}
}

@article{richards1987type,
  title={Type/token ratios: What do they really tell us?},
  author={Richards, Brian},
  journal={Journal of child language},
  volume={14},
  number={2},
  pages={201--209},
  year={1987},
  publisher={Cambridge University Press}
}

@book{smith1967automated,
  title={Automated readability index},
  author={Smith, Edgar A and Senter, RJ},
  volume={66},
  year={1967},
  publisher={Aerospace Medical Research Laboratories, Aerospace Medical Division, Air~…}
}

@article{coleman1975computer,
  title={A computer readability formula designed for machine scoring.},
  author={Coleman, Meri and Liau, Ta Lin},
  journal={Journal of Applied Psychology},
  volume={60},
  number={2},
  pages={283},
  year={1975},
  publisher={American Psychological Association}
}

@article{dale1948formula,
  title={A formula for predicting readability: Instructions},
  author={Dale, Edgar and Chall, Jeanne S},
  journal={Educational research bulletin},
  pages={37--54},
  year={1948},
  publisher={JSTOR}
}

@article{mccannon2019readability,
  title={Readability and research impact},
  author={McCannon, Bryan C},
  journal={Economics Letters},
  volume={180},
  pages={76--79},
  year={2019},
  publisher={Elsevier}
}

@article{gunning1969fog,
  title={The fog index after twenty years},
  author={Gunning, Robert},
  journal={Journal of Business Communication},
  volume={6},
  number={2},
  pages={3--13},
  year={1969},
  publisher={Sage Publications Sage CA: Thousand Oaks, CA}
}

@article{mc1969smog,
  title={SMOG grading-a new readability formula},
  author={Mc Laughlin, G Harry},
  journal={Journal of reading},
  volume={12},
  number={8},
  pages={639--646},
  year={1969},
  publisher={JSTOR}
}

\appendix
\newpage
\onecolumn
\section{Metrics Selection}
\label{app:correlation}

To choose a set of difficulty metrics for our curricula, we began with fifteen candidate metrics spanning six conceptual categories:  
\textit{Information Density} (Compression Ratio),  
\textit{Lexical Diversity} (TTR \cite{richards1987type}, MTLD, HD-D \cite{mccarthy2010mtld}),  
\textit{Readability} (Flesch Reading Ease, Flesch–Kincaid Grade \cite{kincaid1975derivation}, Coleman–Liau Index \cite{coleman1975computer}, automated readability index(ARI) \cite{smith1967automated}, Dale–Chall \cite{dale1948formula}, Linsear Write \cite{mccannon2019readability}, Gunning Fog \cite{gunning1969fog}, SMOG \cite{mc1969smog}),  
\textit{Sequence Length} (Number of Tokens),  
\textit{Fertility} (Fertility Score), and  
\textit{Perplexity} (Perlexity calculated by KenLM \cite{heafield-2011-kenlm})

\begin{figure}[ht]
  \centering
  \includegraphics[width=.8\linewidth]{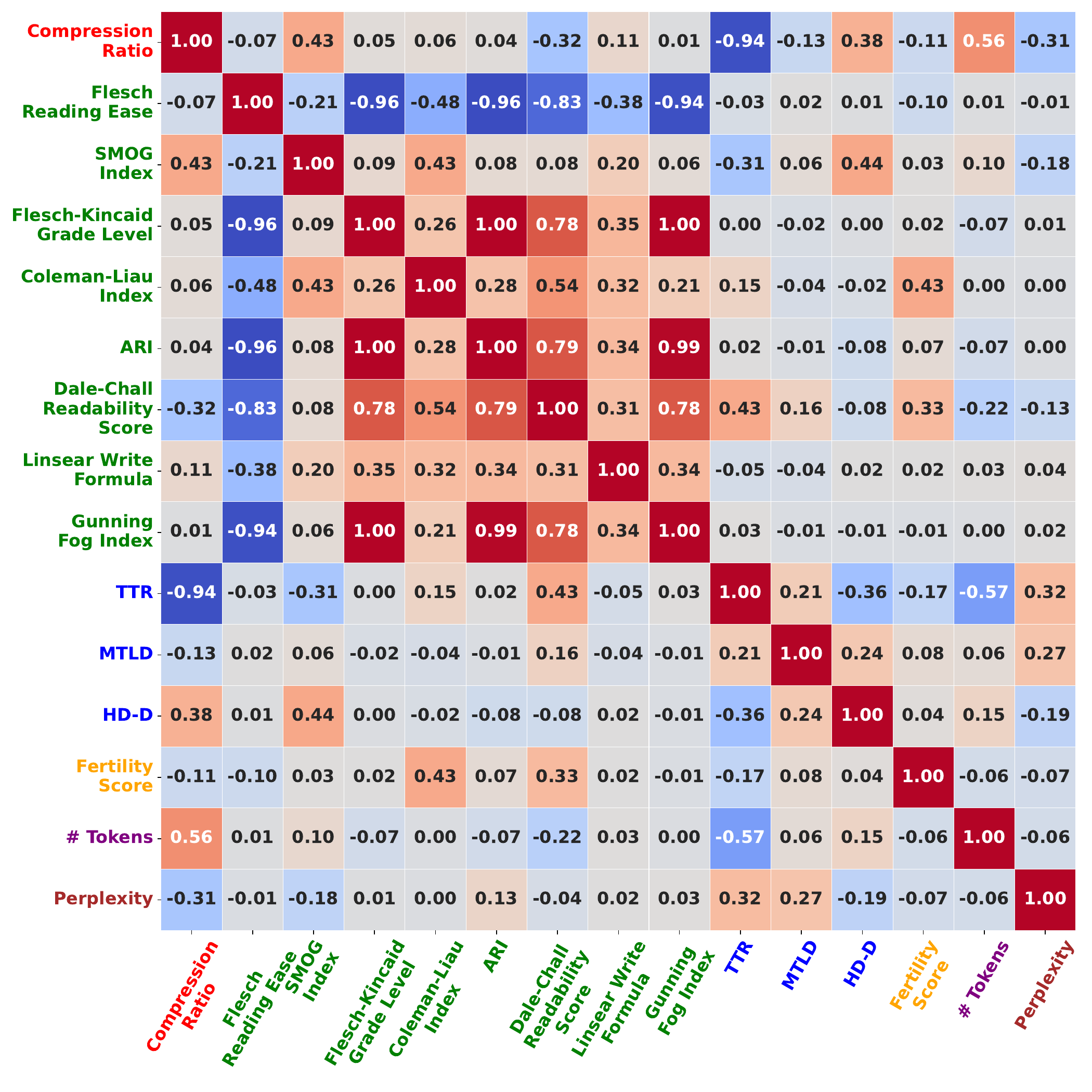}
  \caption{Spearman correlation matrix of the fifteen candidate metrics. Cell shading encodes correlation strength (red = strong positive; blue = strong negative), highlighting the correlations across the different metrics: Density (red), Readability (green), Diversity (blue), Lexical (purple), Fertility (orange), and Perplexity (brown).}
  \label{fig:correlation}
\end{figure}

We computed Spearman rank correlations on our pretraining data to assess monotonic relationships among these metrics.  
Results are presented in Figure~\ref{fig:correlation}.  Strong positive correlations (\(\rho\approx1\)) appear among the various readability formulas, indicating redundancy; moderate correlations (\(\rho\approx0.3\)–0.5) cluster the diversity measures; and near-zero coefficients suggest orthogonality (notably between Perplexity or Fertility and most others).  

Guided by these patterns, we retained one representative from each highly inter-correlated block and preserved metrics that capture unique aspects of text complexity.  Our final six signals are Compression Ratio, Flesch Reading Ease, MTLD, Perplexity, Fertility, and Number of Tokens—ensuring broad coverage of linguistic and information-theoretic difficulty.

\newpage
\section{Curriculum Learning as Warmup}
\label{sec:warmup}
\begin{figure*}[ht]
  \includegraphics[width=0.48\linewidth]{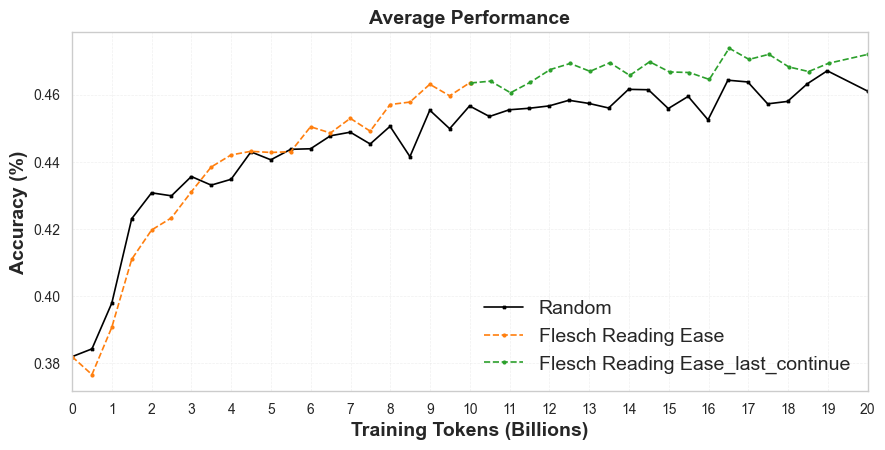} \hfill
  \includegraphics[width=0.48\linewidth]{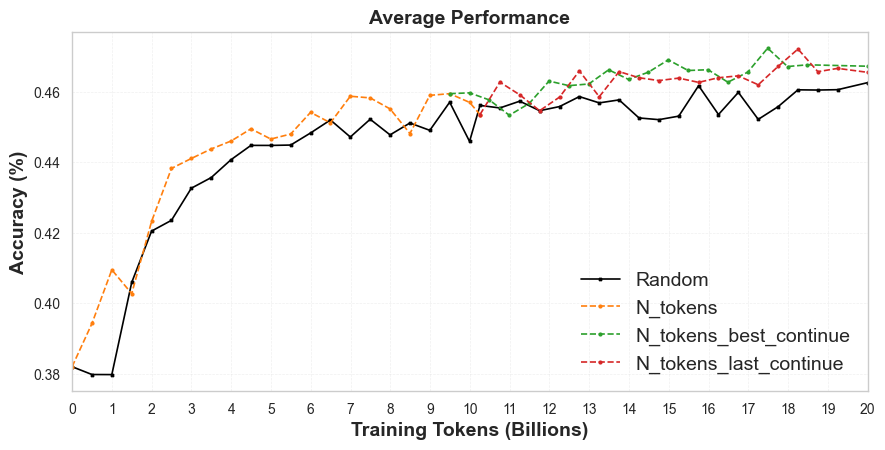}
  \caption {The average accuracy of models continually trained on two curriculum learning settings: Flesch Reading Ease with quadratic pacing (left) and Number of Tokens with vanilla CL (right), we select the best checkpoint and the last checkpoints (they can be the same) from the CL settings to start continual training. }
  \label{fig:ablwarmup}
\end{figure*}

In Figure \ref{fig:ablwarmup} we show two more experiments that use curriculum learning as warmup: the Flesch Reading Ease with quadratic pacing and Number of Tokens with vanilla curriculum learning. The models trained on both settings and their baselines are then continually trained on randomly shuffled data.

For the model trained on Flesch Reading Ease with quadratic pacing setting, we continue training the model from the last checkpoint since it is also the best. We observed a similar pattern as in Figure \ref{fig:mtld_continue}: CL as warmup setting consistently outperforms the baseline by a large margin, especially in the continual training phase. Specifically, this setting achieves $1.5\%$ higher accuracy and reaches the best baseline performance with $40.0\%$ fewer steps.

For the model trained on Number of Tokens with vanilla CL setting, we chose both the best and last checkpoint to start continual training. The models trained from both checkpoints show close performance throughout training, with the model starting from the best checkpoint slightly better than the model starting from the last checkpoint. Overall the CL as warmup settings achieve $2.2\%$ higher accuracy than the baseline and reached the best baseline performance with $20.7\%$ fewer steps.

\section{Number of difficulty groups}
\label{sec:numgroups}

\begin{figure*}[!h]
  \centering
  \includegraphics[width=\linewidth]{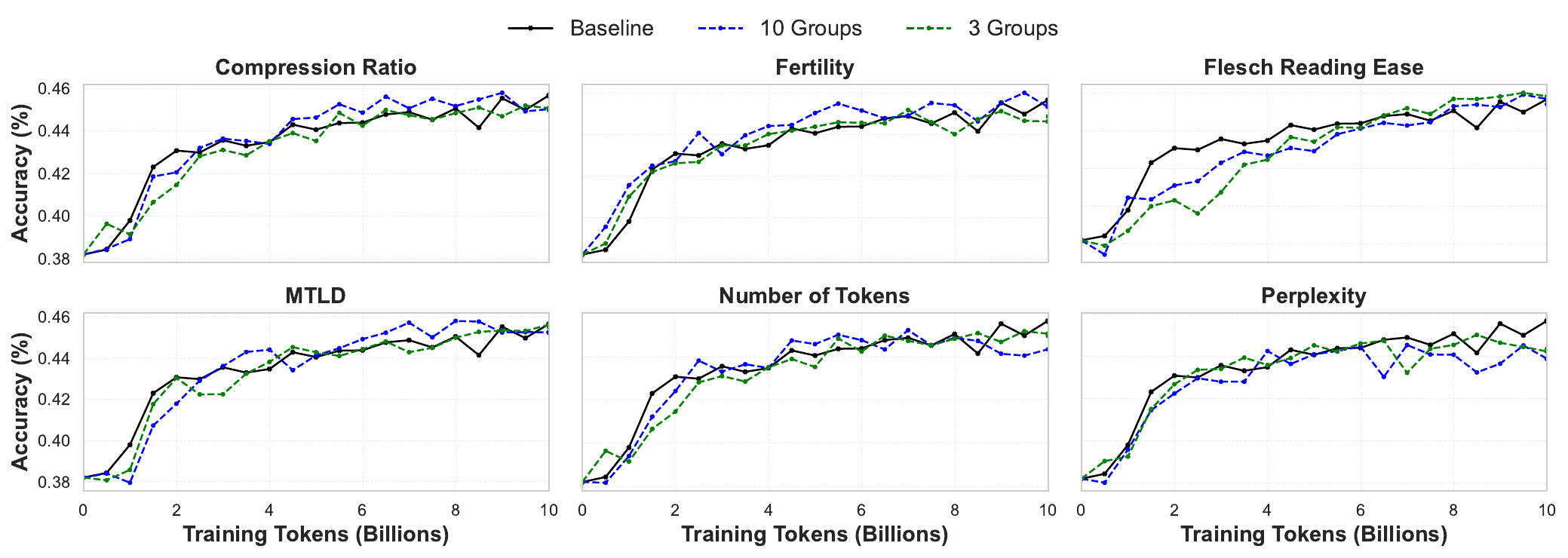}
  \caption{The comparison of the average accuracy for models trained via CL on 6 difficulty metrics under the linear pacing function with 3 and 10 difficulty groups.}
  \label{fig:group10vs3}
\end{figure*}

We investigate the influence of using a different number of difficulty groups, and show the results when using 3 difficulty groups compared with our default 10 difficulty groups with the linear pacing CL setting in Figure \ref{fig:group10vs3}. Though 3 difficulty groups can be seen as easy, middle and hard groups intuitively, the performance is generally worse or similar to 10 groups with a small difference. We hypothesize that 3 groups lead to coarser transitions between groups and potentially lead to model overfitting to each group, which can be harmful for model training. But overall, we consider the result is less sensitive to the choice of the number of groups.

\section{Robustness of Curriculum Learning}
\label{sec:appendix2}
\begin{figure*}[!h]
  \centering
  \includegraphics[width=\linewidth]{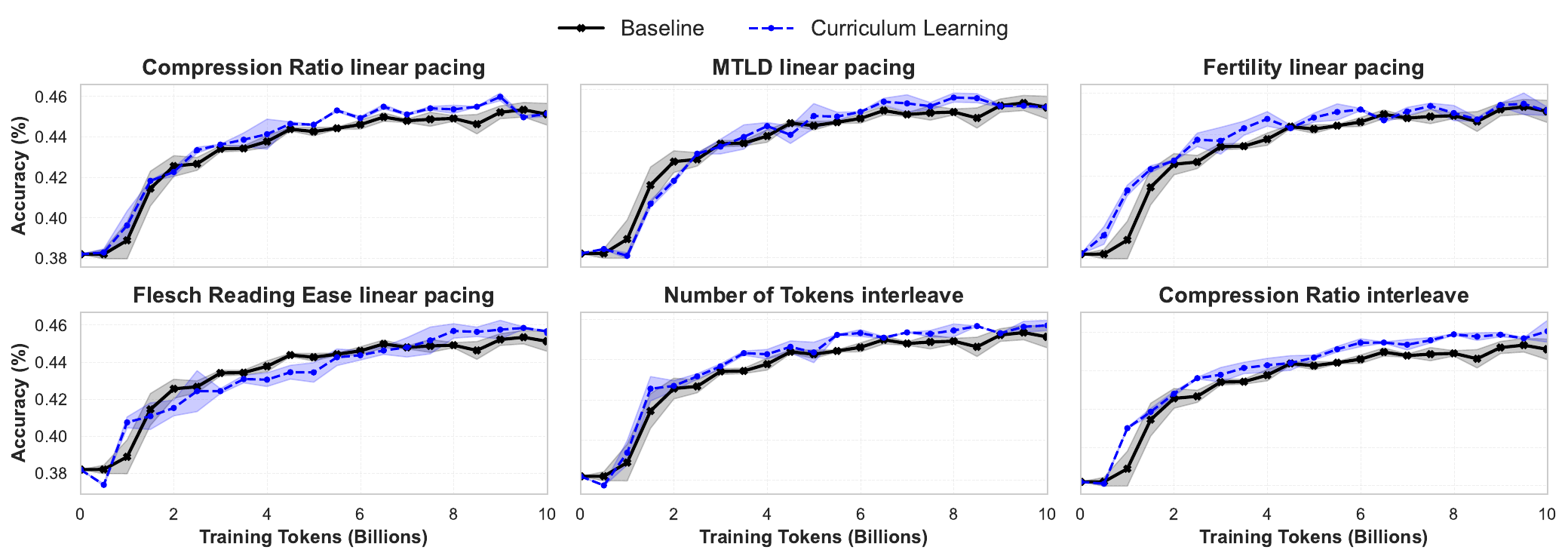}
  \caption{Performance comparisons between baseline (black) and curriculum learning (blue) across six high-performing settings. Accuracy is averaged over three runs, with shaded areas denoting standard deviations. Results confirm the robustness of curriculum learning, with consistent improvements and minimal overlap in variation with the baseline.}
  \label{fig:robust}
\end{figure*}

We select 6 settings that previously showed strong performance: Compression Ratio, MTLD, Fertility and Flesch Reading Ease with linear pacing; Number of Tokens and Compression Ratio with interleave pacing, and we repeat two more times for each setting, as well as the baseline. Figure \ref{fig:robust} shows the results with average performance across runs and the standard deviations. For all tested settings, the previously established findings still hold on average, even considering the standard deviation, as there are very small part of overlapping between the variation range for CL settings and the baseline. This shows the robustness of our defined CL settings.

\section{Sampling vs Ordering}
\label{sec:samplevsorder}    
\begin{figure*}[!h]
  \centering
  \includegraphics[width=\linewidth]{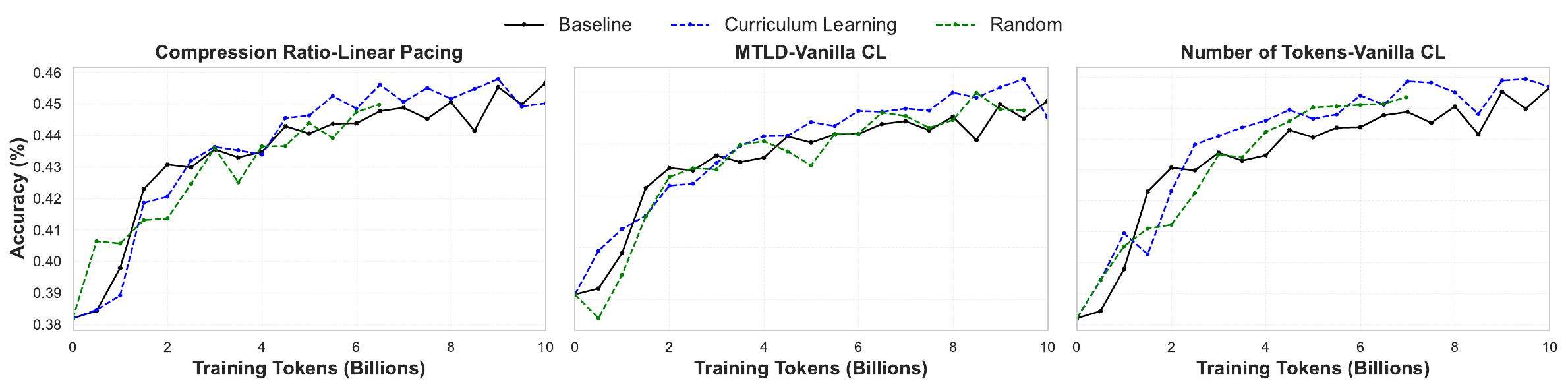}
  \caption{Performance comparisons between curriculum learning ordering and random shuffling of the data selected up to a selected difficulty level. Though leaving out the most difficult samples can be viewed as an effective data cleaning process, the performance is still lower than the ordered setting using the same data.}
  \label{fig:sample_average_grid}
\end{figure*}
We conduct a controlled comparison to disentangle the effects of data selection from data ordering. We select 3 CL settings where the model reached best performance early but experience a performance drop at last steps: MTLD, Number of Tokens with vanilla curriculum learning and Compression Ratio with linear pacing. For each, we extract the training data used before the best-performing checkpoint, then randomly shuffle this subset and retrain the model. While these shuffled subsets outperform the full random baseline—suggesting a filtering effect—the ordered curriculum versions still achieve superior performance. This indicates that the benefits of curriculum learning arise not only from selecting informative data but also from the order in which it is presented during training.

\section{Cerebras}
\label{sec:cerebras}

CS-2 systems are purpose-built network-attached AI accelerators. Each CS-2 features 40~GB of SRAM and a peak of 62.5 AI PetaFLOPs, providing a total of 4 ExaFLOPs of AI compute across 64 systems in the CG-2 supercomputer. Utilizing the weight streaming mode of the Cerebras software stack, the Condor Galaxy supercomputers can flexibly schedule multiple jobs based on hardware resource requirements and priority. The number of CS-2s allocated to a job can be dynamically adjusted during training, with performance scaling linearly up to 64 CS-2s per job. This scalability is facilitated by the Cerebras software stack’s use of pure data parallelism to distribute the workload across multiple CS-2s. Jobs are managed by a priority queue system, ensuring efficient allocation of computational resources.

\medskip

MemoryX is a large-capacity off-wafer memory service used to store all model weights, gradients, and optimizer states. SwarmX is a broadcast/reduce fabric that connects the memory service MemoryX to each of the CS-2 systems in a wafer-scale cluster. SwarmX coordinates the broadcast of the model layer weights, giving each CS-2 a local copy, and it receives and aggregates (by addition) the independent weight gradients coming from the CS-2 systems during backpropagation. At the end of each iteration, the aggregated gradients are sent to MemoryX for weight update.

\subsection{Reproducibility on Nvidia GPUs}
To further eliminate the difference between models trained on the Cerebras System and Nvidia GPUs, we trained two 0.5B models with 10B tokens using Cerebras CS-2 System and Nvidia A100 GPU respectively, and evaluate the model performance on eight benchmarks, as shown in Table \ref{tab:perf-diff}, we report the evaluation result for both models and the relative difference compared with model trained on Cerebras. Across all benchmarks, the relative difference is less than $0.5\%$, so we consider the model trained on both hardware makes no difference regarding model performance.

\begin{table}[!h]
\centering
\small
\begin{tabular}{lccc}
\hline
\textbf{Benchmark} & \textbf{Cerebras} & \textbf{NVIDIA A100} & \textbf{$\Delta$\% vs Cerebras} \\
\hline
ArcChallenge & 0.1928 & 0.1932 & 0.21 \\
BoolQ        & 0.6101 & 0.6125 & 0.39 \\
COPA         & 0.6900 & 0.6902 & 0.03 \\
Hellaswag    & 0.3087 & 0.3092 & 0.16 \\
OpenBookQA   & 0.1820 & 0.1823 & 0.16 \\
PIQA         & 0.6561 & 0.6572 & 0.17 \\
WinoGrande   & 0.4893 & 0.4903 & 0.20 \\
xwinograd\_en & 0.5939 & 0.5948 & 0.15 \\
\hline
\end{tabular}
\caption{Performance comparison of NVIDIA A100 relative to Cerebras across multiple benchmarks.}
\label{tab:perf-diff}
\end{table}


\section{Scale Curriculum Learning}
\label{sec:scale}

We evaluate the scalability of our findings by replicating key curriculum learning settings using 1B and 3B models following the Llama3.2 configuration. The training hyperparameters are the same as previous experiments.

\subsection{Vanilla Curriculum Learning}
\begin{figure*}[!h]
  \centering
  \includegraphics[width=\linewidth]{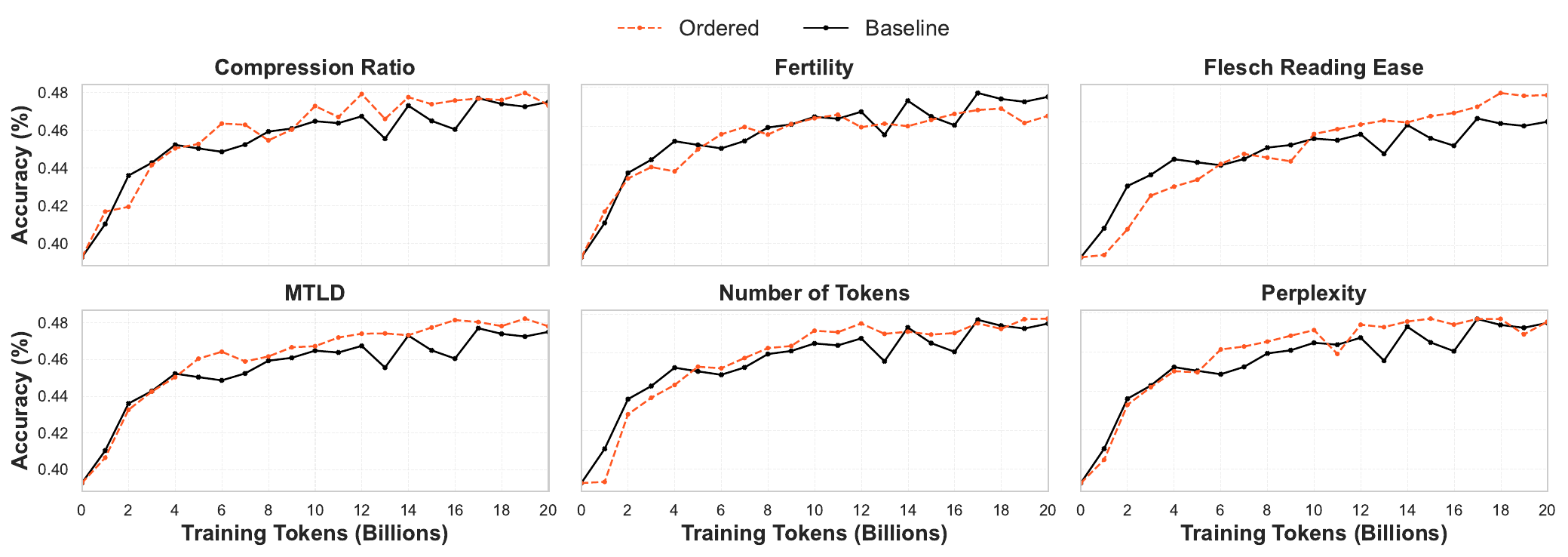}
  \caption{Performance of vanilla curriculum learning at scale using 1B model trained on 20B tokens compared to the baseline. Most settings show a faster improvement in the early to mid stage, with MTLD and Flesch Reading Ease showing a strong performance.}
  \label{fig:3p21bvanilla}
\end{figure*}
In this section, we experiment with the vanilla CL setting, we use the same set of difficulty metrics and instead of selecting a dataset of 10B, we select 20B tokens according to the scaling law, and create 6 strictly ordered training set as well as a baseline set, we train a model for each set.

Figure \ref{fig:3p21bvanilla} shows the results. We observe similar findings to our previous experiments: (1) the CL settings and the baseline differentiate the most in the early to mid training stages, with most CL settings reaching the best performance faster than the baseline; (2) the MTLD setting consistently outperforms the baseline during the training; (3) Number of Tokens setting has a faster performance improvement in the early and mid stage of training; (4) Flesch Reading Ease setting keeps a constant improvement rate of performance, interestingly, outperforms the baseline by a large margin at the end of training.

\subsection{Flesch Reading Ease with quadratic pacing}
Since the Flesch Reading Ease setting showed promising and unique results from all previous settings: a constant rate of model performance improvement and didn't show a sign of convergence. We want to test this specifically for a larger model, thus, we chose the setting that shows the biggest improvement rate and final performance-quadratic pacing-to train a 1B model.

\begin{figure*}[!h]
  \centering
  \includegraphics[width=.7\linewidth]{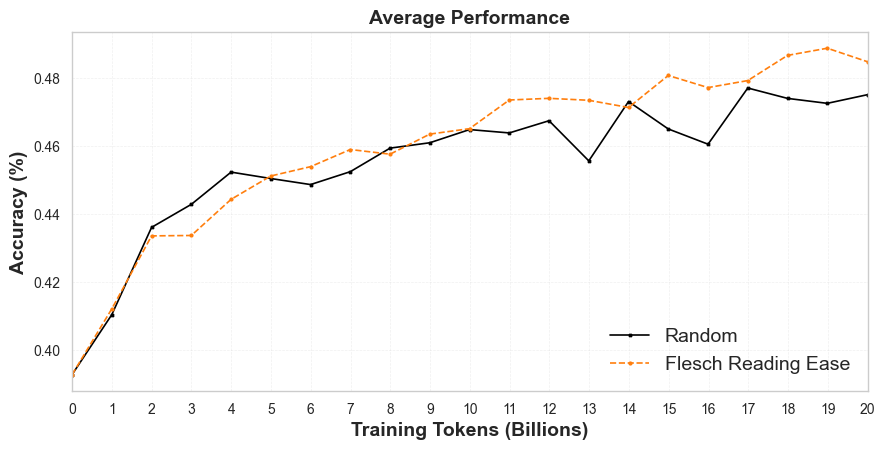}
  \caption{Performance of using Flesch Reading Ease with quadratic pacing, the model shows stable, non-convergent performance improvements across the full training horizon.}
  \label{fig:3p21bfrequa}
\end{figure*}

As shown in Figure \ref{fig:3p21bfrequa}, we observed again a stable performance gain till the end of the training for Flesch Reading Ease with quadratic pacing setting, and still no sign of convergence. The model trained on this setting achieves $2.5\%$ higher accuracy, showing again the potential of this difficulty metric.
\newpage
\subsection{Curriculum Learning as Warmup}
We are also interested in applying CL under S\ref{hyp:third} that use CL as a warmup phase of training, which previously showed impressive results.

\begin{figure*}[ht]
  \includegraphics[width=0.48\linewidth]{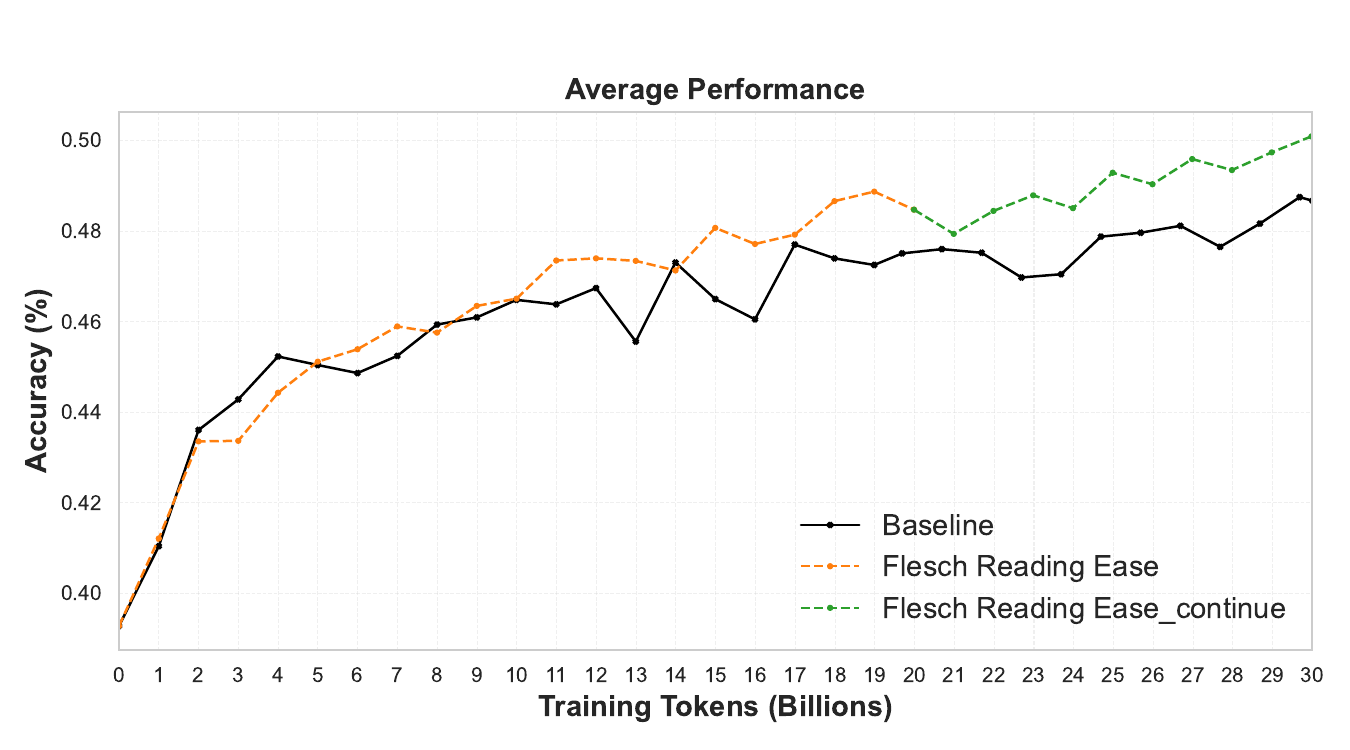} \hfill
  \includegraphics[width=0.48\linewidth]{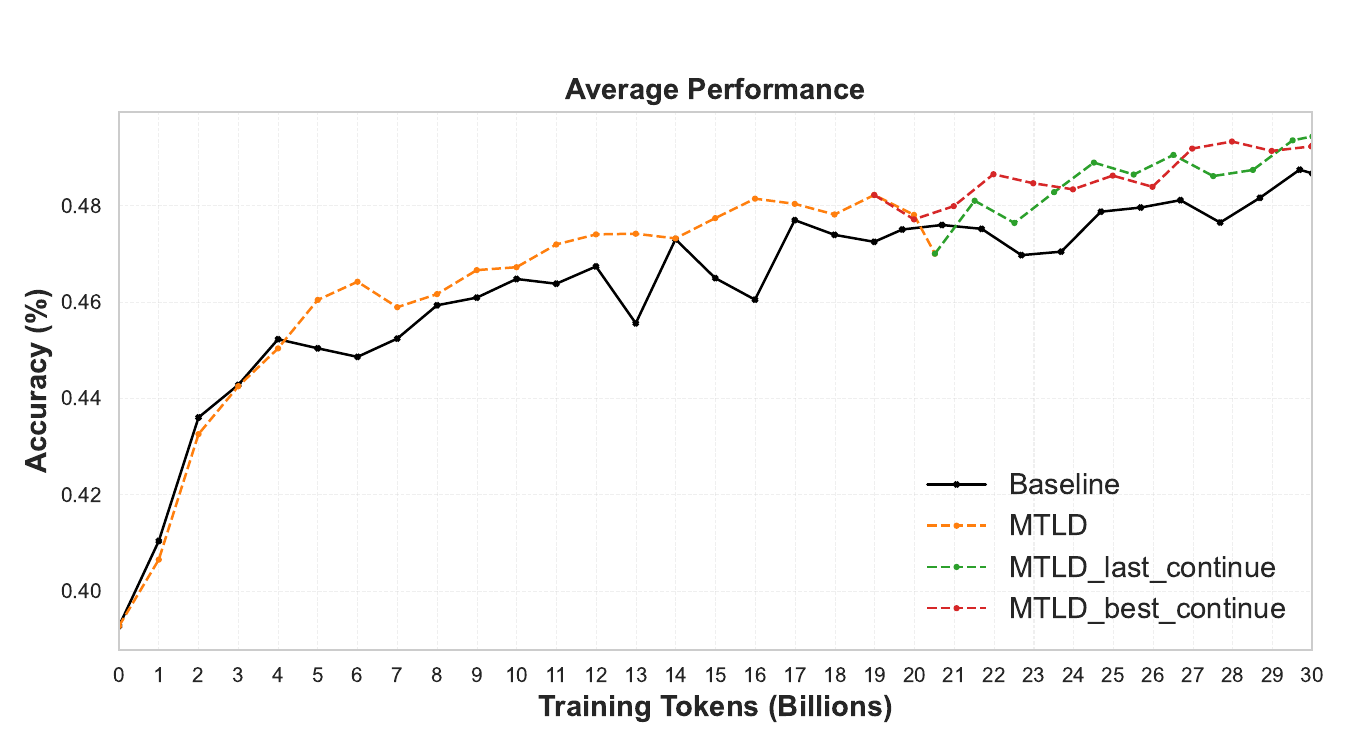}
  \caption {The average accuracy of models continually trained on two curriculum learning settings: Flesch Reading Ease with quadratic pacing (left) and MTLD with vanilla CL (right), the models were trained using 20B tokens for CL warmup, then we select the best checkpoint and the last checkpoints (they can be the same) from the CL settings to start continual training for the next 10B tokens. }
  \label{fig:abl3p21bwarmup}
\end{figure*}

We tested two settings: Flesch Reading Ease with quadratic pacing and MTLD with vanilla CL to serve as CL warmup pretraining with 20B tokens, then we continue training the model from either the best or the last chekcpoint for 10B tokens which are randomly shuffled. Our results is shown in Figure \ref{fig:abl3p21bwarmup}, where we observe same pattern as the 0.5B model: CL warmup lead the model achieves higher performance throughout the training, especially the continual training phase. Specifically, the Flesch Reading Ease setting and the MTLD setting achieved $3.2\%$ and $2.1\%$ higher peak performance compared to the baseline.

\begin{figure*}[!h]
  \centering
  \includegraphics[width=0.7\linewidth]{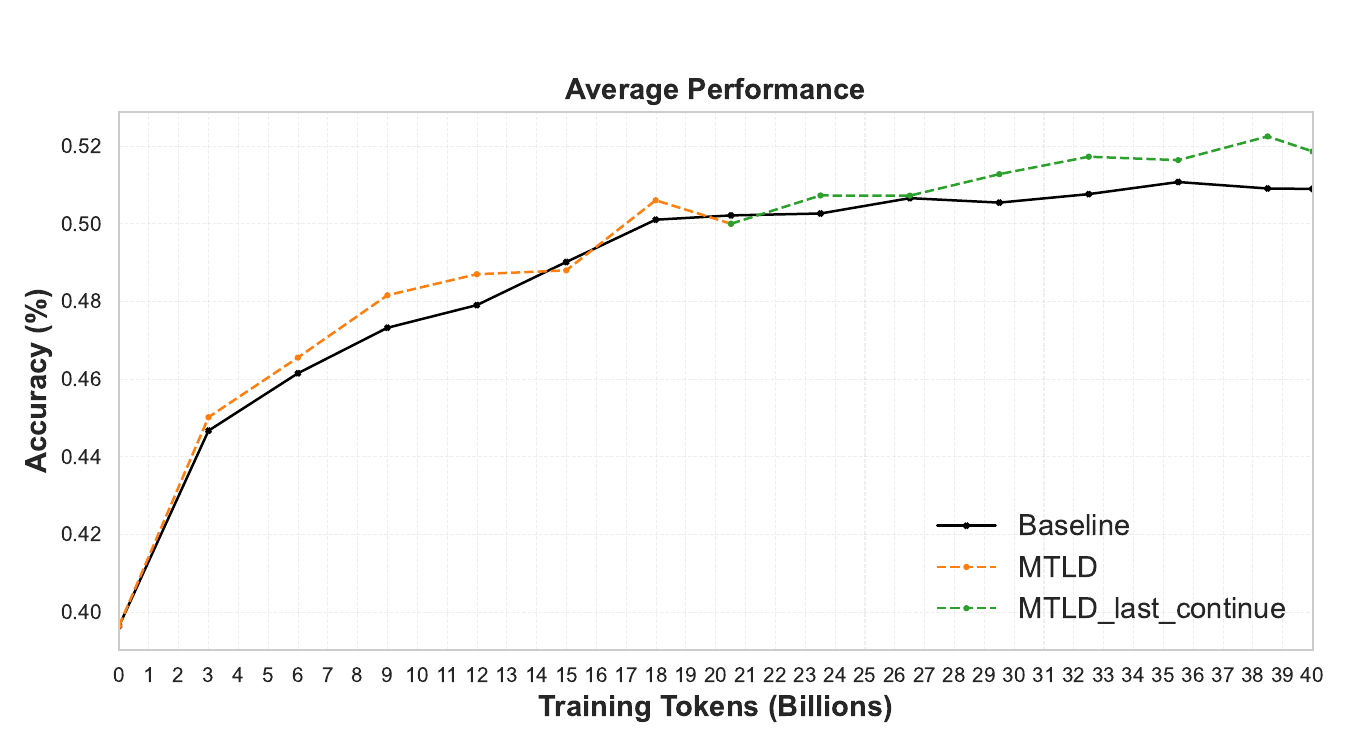}
  \caption {The average accuracy of model continually trained on MTLD with vanilla CL, the model is trained using 20B tokens for CL warmup, then we select the last checkpoint from the CL settings to start continual training for the next 20B tokens. }
  \label{fig:abl3p23bwarmup}
\end{figure*}

We then tested a 3B model on the MTLD setting with the same setup as in Figure \ref{fig:abl3p23bwarmup}. We first train the model under curriculum learning for 20B tokens, then perform random training with 20B more tokens.

For the 3B model, we observed the same finding as in 1B and 0.5B models - CL warmup consistently improves performance, achieving a 2.0\% gain over the baseline. These findings showed the scalability of CL as warmup setting, where the model can achieve consistent performance gains compared to the random baseline.

\begin{figure*}[!h]
  \centering
  \includegraphics[width=0.7\linewidth]{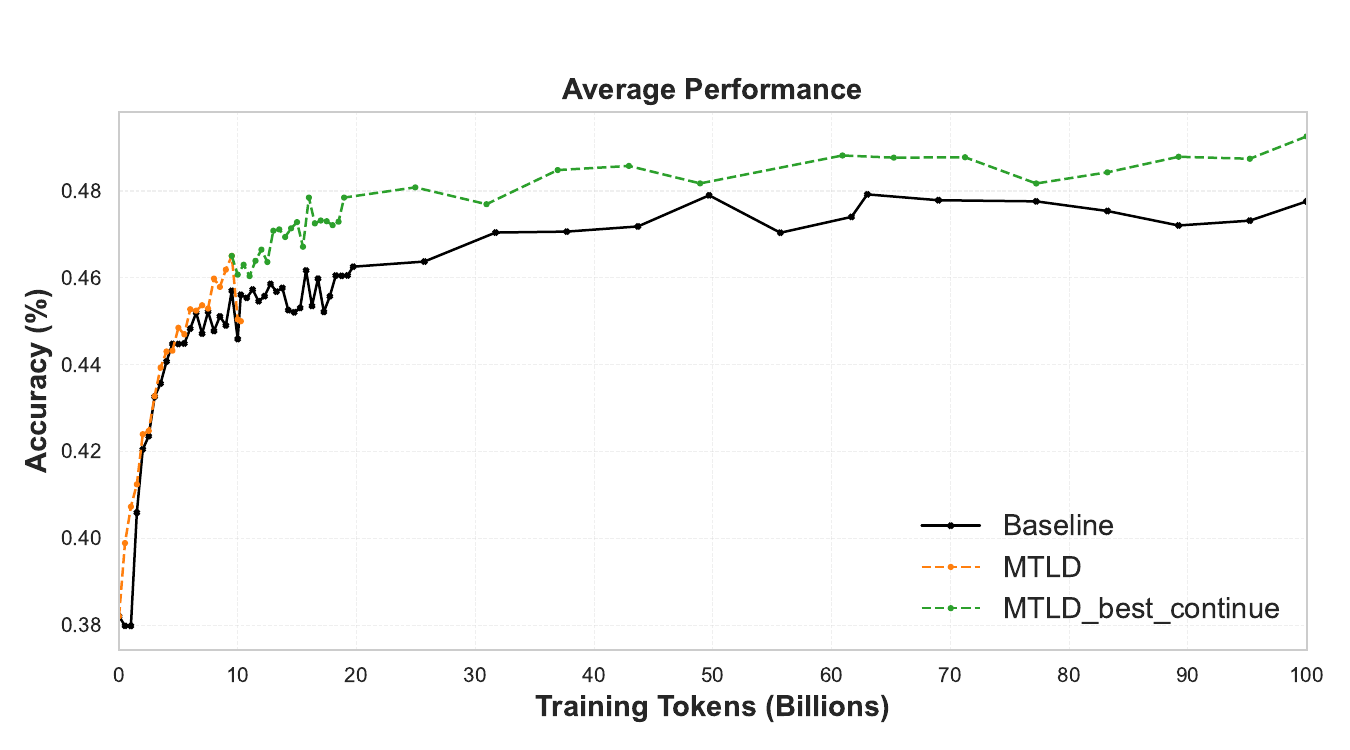}
  \caption {The average accuracy of the model trained under CL as warmup setting using MTLD as the difficulty metric, extended to 100B training tokens.}
  \label{fig:mtld100b}
\end{figure*}

To further extend our main setting, we apply CL as a warmup to a 100B-token training regime, a scale at which prior work has observed the emergence of new model capabilities. We initialize training from the MTLD\_best checkpoint in Figure \ref{fig:mtld_continue}. Both the CL and baseline models are then trained to 100B tokens under an identical training setup as used throughout the paper. As shown in Figure \ref{fig:mtld100b}, CL as warmup consistently outperforms the random baseline across all checkpoints. At 100B tokens, the warmup model achieves an absolute performance improvement of approximately 3.1\%, demonstrating the robustness and consistency of our findings at larger training scales.

\section{Computation overhead}
\label{sec:overhead}
One important aspect of designing efficient curriculum learning settings is the computation overhead of computing the difficulty metrics. We show the gains in our experiments come without sustained additional cost—CL involves only restructuring the initial data phase, making it a practical and low-overhead approach. We provide Table \ref{tab:metrics} showing time (rounded to 0.5 min) and cost estimates on a 72 vCPU AWS machine for computing difficulty metrics over 10B tokens. Additional cost for data loading and data ordering is ~2\$. While training a 0.5B model with 10B tokens requires ~3e19FLOPs, accounting for about 84 GPU hours of A100. The computation overhead for calculating data difficulty and ordering data is thus negligible compared to pretraining.

\begin{table}[!h]
\centering
\small
\begin{tabular}{lcccccc}
\hline
\textbf{Metric} & \textbf{Copression Ratio} & \textbf{Fertility} & \textbf{Readability(FRE)} & \textbf{MTLD} & \textbf{\#Tokens} & \textbf{Perplexity} \\
\hline
Time(mins)/cost(\$) & 11.5/0.7 & 29.4/1.8 & 21/1.3 & 23.6/1.4 & 19.5/1.2 & 13/0.8 \\
\hline
\end{tabular}
\caption{Computation time and cost for the difficulty metrics.}
\label{tab:metrics}
\end{table}

\end{document}